\documentclass[11pt]{article}

\usepackage[preprint]{acl}

\usepackage{graphicx}	
\usepackage{amsmath}	
\usepackage{amssymb}	
\usepackage{booktabs}
\usepackage{times}
\usepackage{microtype}
\usepackage{epsfig}
\usepackage{caption}
\usepackage{float}
\usepackage{placeins}
\usepackage{color, colortbl}
\usepackage{stfloats}
\usepackage{enumitem}
\usepackage{tabularx}
\usepackage{xstring}
\usepackage{multirow}
\usepackage{xspace}
\usepackage{subcaption}
\usepackage{xcolor}
\usepackage[hang,flushmargin]{footmisc}
\usepackage{makecell}

% ...

\definecolor{myorange}{HTML}{FF6700}
\definecolor{mypurple}{HTML}{9900ff}
\definecolor{mygreen}{HTML}{009000}

\newcommand{\model}{v-\textsc{Hub}\xspace}
\newcommand{\fulltitle}{A Benchmark for Video Humor Understanding \\from Vision and Sound\xspace}

\def\dataset{\model}

\usepackage{times}
\usepackage{latexsym}
\usepackage{cleveref}

\usepackage{marvosym}

\usepackage[T1]{fontenc}

\usepackage[utf8]{inputenc}

\usepackage{microtype}

\usepackage{inconsolata}

\usepackage{graphicx}

\newcommand{\blfootnote}[1]{%
  \begingroup
  \renewcommand\thefootnote{}\footnote{#1}%
  \addtocounter{footnote}{-1}%
  \endgroup
}

\title{\model: \fulltitle}

\author{
\begin{tabular}{@{}l@{}}
\textbf{Zhengpeng Shi}$^{1,3}$ \quad\,\,\,\,
\textbf{Yanpeng Zhao}$^{3}\,^{\dagger}\,\textsuperscript{\Letter}$ \quad\,\,\,\,
\textbf{Jianqun Zhou}$^{2,3}$ \quad\,\,\,\,
\textbf{Yuxuan Wang}$^{4}$ \\
\textbf{Qinrong Cui}$^{4}$ \quad\,\,\,\,
\textbf{Wei Bi}$^{4}$ \quad\,\,\,\,
\textbf{Songchun Zhu}$^{3}$ \quad\,\,\,\,
\textbf{Bo Zhao}$^{1}\,\textsuperscript{\Letter}$ \quad\,\,\,\,
\textbf{Zilong Zheng}$^{3}\,\textsuperscript{\Letter}$
\end{tabular}
\\[0.35em]
\begin{tabular}{@{}c@{}}
$^{1}$Shanghai Jiao Tong University \quad
$^{2}$Wuhan University \\
$^{3}$Beijing Institute for General Artificial Intelligence \\
$^{4}$Independent Researcher 
\end{tabular}
}

\newcommand{\checkicon}{\raisebox{-.25em}{\includegraphics[width=1em]{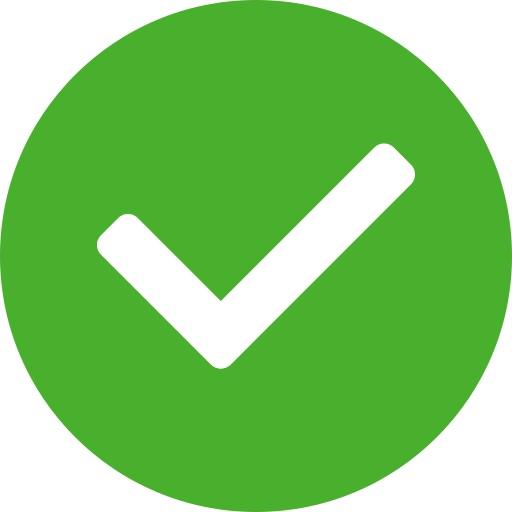}}}
\newcommand{\crossicon}{\raisebox{-.25em}{\includegraphics[width=1em]{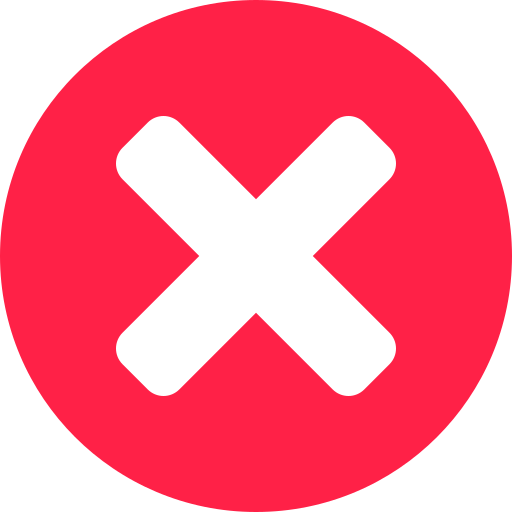}}}

\begin{document}

\maketitle

\blfootnote{
\resizebox{\linewidth}{!}{%
\begin{tabular}{@{}l@{}}
$\dagger$: Project Lead. \ \Letter: Corresponding Author. \\
Contact: \texttt{shi\_zpeng@sjtu.edu.cn,yannzhao.ed@gmail.com}.
\end{tabular}
}
}

\begin{abstract}
AI models capable of comprehending humor hold real-world promise---for example, enhancing engagement in human-machine interactions. To gauge and diagnose the capacity of multimodal large language models (MLLMs) for humor understanding, we introduce \dataset, a novel video humor understanding benchmark. \dataset comprises a curated collection of non-verbal short videos, 
reflecting real-world scenarios where humor can be appreciated purely through visual cues. We pair each video clip with rich annotations to support a variety of evaluation tasks and analyses, including a novel study of environmental sound that can enhance humor.
To broaden its applicability, we construct an open-ended QA task, making \dataset readily integrable into existing video understanding task suites. We evaluate a diverse set of MLLMs, from specialized Video-LLMs to versatile OmniLLMs that can natively process audio, covering both open-source and proprietary domains. The experimental results expose the difficulties  MLLMs face in comprehending humor from visual cues alone. 
Our findings also demonstrate that incorporating audio helps with video humor understanding, highlighting 
the promise of integrating richer modalities for complex video understanding tasks.


\end{abstract}

\begin{figure}[ht!]
\begin{subfigure}[t]{\linewidth}
    \centering
    \includegraphics[width=\linewidth]{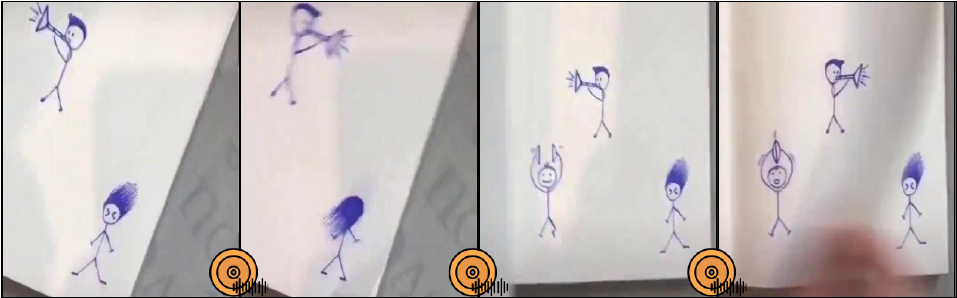}
    \caption[teaser-c]%
    {{\footnotesize \textbf{Visuals}+Audio. As the man flips through the pages, \textcolor{mygreen}{cartoon characters gradually appear}, accompanied by a distinct melody. First, the dancer's rhythm and the \textcolor{mygreen}{suona player's piercing tune, then the cymbal player's resonant clash}, together creating \textcolor{black}{an evolving effect}.
    }}    
    \label{fig:teaser-c}
\end{subfigure}
\begin{subfigure}[t]{\linewidth}
    \centering
    \includegraphics[width=\linewidth]{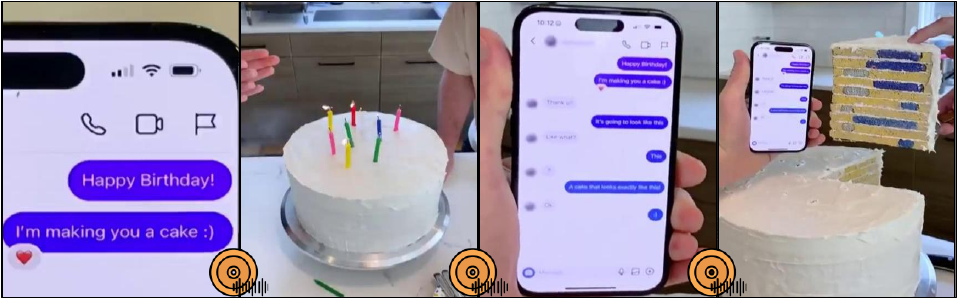}
    \caption[teaser-d]%
    {{\footnotesize \textbf{Visuals}+Audio+Text.
    \textcolor{mygreen}{A guy messaged his friend} that he was making a birthday cake for them. After it was baked and sliced, \textcolor{mygreen}{the inside mimicked their chat bubble layout}. The whole scene was made merrier by \textcolor{mygreen}{the Happy Birthday melody}.
    }}    
    \label{fig:teaser-d}
\end{subfigure}
\caption{\label{fig:teaser-small}
Examples of visual-centric humor understanding, where `audio' and `text' refer to environmental sound (\emph{cf.} human speech) and visual text, respectively.
}
\end{figure}

\begin{table*}[]
\centering
\caption{\textcolor{black}{
Humor understanding benchmark comparison.
Modality indicates the modalities used for humor understanding, where T, V, and A denote text, visual, and audio, respectively. Our \dataset is \emph{visual-centric}, as it contains humor derived predominantly from visual cues (V) and enhanced by environmental sound (A).
}}
\label{tab:dataset_comparison}
\resizebox{1\linewidth}{!}{%
\begin{tabular}{@{}c|c|c|c|c|ccc@{}}
\toprule
\multirow{2}{*}{\textbf{Dataset}}                         & \multirow{2}{*}{\textbf{Type}} & \multirow{2}{*}{\textbf{Source}}           & \multirow{2}{*}{\makecell{\textbf{Average} \\ \textbf{Duration}}} & \multirow{2}{*}{\textbf{Modality}} & \multicolumn{3}{c}{\textbf{Tasks}}     \\ \cmidrule(l){6-8} 
                                                          &                                &                                   &                               &                                                  & Explanation & Matching & Open-ended QA \\ \midrule
NYCC \citep{hessel2022androids}          & Cartoon                        & New Yorker Caption Contests       & ---                            & T, V\hphantom{, A}                                             & \checkicon & \checkicon & \crossicon               \\
Oogiri-GO \citep{zhong2024let}  & Images        & Oogiri Game   & --- & T, V\hphantom{, A}                                            & \crossicon & \checkicon & \crossicon             \\
MUStARD \citep{castro-etal-2019-towards} & Sitcom                         & TV Shows                          & 19 sec.                    & T, V, A                                             & \crossicon & \checkicon & \crossicon               \\
WITS \citep{kumar-etal-2022-become}      & Sitcom                         & TV Shows                          &  17 sec.                            & T, V, A                                             & \crossicon & \checkicon & \crossicon               \\
UR-FUNNY \citep{hasan-etal-2019-ur}             & Speech                         & TED Talks                         & 20 sec.                    & T, V, A                                            & \crossicon & \checkicon & \crossicon              \\
SMILE \citep{hyun-etal-2024-smile}              & Sitcom, Speech                 & TV Shows, TED Talks               & 28 sec.                  & T, V, A                                             & \checkicon & \crossicon & \crossicon              \\
ExFunTube \citep{ko2023can}              & Short videos                   & Youtube                           &  16 sec.                            & T, V, A                                            & \checkicon & \crossicon & \crossicon             \\
HumorQA \citep{xie2024funqa}               & Surprising videos              & Youtube                           & \hphantom{0}7 sec.                     & T, V, A                                            & \checkicon & \checkicon & \checkicon               \\ \midrule
\textbf{\dataset} (ours)                            & \makecell{Short videos, \\\textbf{Silent comedies} }    & \makecell{X, Youtube, \\Charlie Chaplin’s \textbf{silent films}} & \textbf{14} sec.                    & \hphantom{T,} V, A                                                & \checkicon & \checkicon & \checkicon              \\ \bottomrule
\end{tabular}
}

\vspace{-1em}
\end{table*}

\section{Introduction}\label{sec:intro}

Humor enriches our daily lives and appears in many forms, from jokes and cartoons to comedies and viral videos. AI models capable of understanding humor hold promise for engaging with humans empathetically~\citep{Hampes_2001,Hampes_2010}, but perceiving and comprehending humor can be challenging even to humans due to the heavy reliance on nontrivial reasoning, social and cultural contexts, etc (see Figure~\ref{fig:teaser-small}). This, on the other hand, makes humor understanding a promising testbed to evaluate how well state-of-the-art AI models understand humor. Indeed, there has been a line of research centering around gauging the capability of pre-trained large language models (LLMs) for humor understanding from visuals and text~\citep{hessel2022androids, hyun-etal-2024-smile, ko2023can, chen2024talk}, but parallel studies of \emph{multimodal} LLMs are limited, though they are naturally suited for understanding multimodal humor.

In this work, we address this gap by investigating humor understanding with multimodal LLMs (MLLMs), focusing specifically on MLLMs that can process video. We choose video as the primary medium of humor, since it captures nuanced variations and diverse styles, presenting a unique challenge. For example, perceiving the humor in Figure\ref{fig:teaser-d} requires recognizing visual text (or background music) and the layout of chat bubbles and understanding their temporal and semantic correspondences with the cut surface of the cake slice. 

While there have been a few benchmarks on video humor understanding (see Table~\ref{tab:dataset_comparison}), most of them are designed exclusively for the evaluation of LLMs~\citep{ko2023can, hyun-etal-2024-smile},\footnote{They translated videos into language descriptions and performed verbal humor evaluation with LLMs.} and the curated humor is dominated by linguistic cues,
ignoring the fact that humans can understand humor from visual cues alone, exemplified by silent comedies. Though \citet{xie2024funqa} have conducted a humor understanding evaluation of MLLMs, the evaluation is limited in coverage and temporal complexity. Moreover, as an important component of non-verbal humor, environmental sound has been overlooked in their data curation and evaluation. 


To address these limitations, we curate a set of visual-centric humorous videos from two complementary sources: Charlie Chaplin’s silent films and user-generated short funny videos. Silent film humor is conveyed through visual cues, but is thematically and culturally constrained due to the scripted performance. To increase diversity, we incorporate user-generated funny short videos from various occasions and cultural backgrounds. We rigorously filtered the videos to retain only those where the humor has no reliance on speech and has a duration longer than 5 seconds. Our final dataset consists of videos where humor is derived predominantly from the visual modality, making it visual-centric and better suited for evaluating MLLMs, including variants that can natively process audio.\footnote{In this work, audio primarily refers to environmental sound rather than human speech (see Section~\ref{sec:data-filter}).}

To assess how well MLLMs understand humor in video, we create \dataset, a video humor understanding benchmark.
\dataset offers two typical evaluation tasks for humor understanding.
(1) First, the \emph{Caption Matching} task challenges MLLMs to align video captions with the corresponding videos. 
(2) Second, the \emph{Humor Explanation} task evaluates whether MLLMs can extract humor elements and provide accurate rationales. 
To broaden the applicability of \dataset, we further construct (3) an
\emph{Open-ended QA} task that evaluates the MLLMs’ fundamental understanding of videos from the humor genre across temporal, descriptive, and causal dimensions.
Together, these tasks provide a comprehensive framework to benchmark MLLMs in visual-centric humor understanding.

We evaluate representative MLLMs from both open- and closed-source domains. Depending on the input modalities, we consider the following three task settings.
(1) The \textit{Text-Only} setting assumes human-level interpretation of video contents and provides detailed human-written descriptions.
(2) The \textit{Video-Only} setting offers only videos (without audio) to assess the ability of MLLMs to derive humor solely from visual cues.
(3) We further propose a novel \textit{Video+Audio} setting that combines visual and auditory signals to determine whether sound cues—such as background music and sound effects—help MLLMs (aka. OmniLLMs) better understand humor.

We empirically find that MLLMs generally perform better with text-only inputs than with video-only inputs.
For example, Qwen2.5-VL drops in accuracy from 0.726 to 0.666 on \emph{Caption Matching},
indicating its struggles in capturing subtle visual cues for humor understanding. Adding audio yields notable improvements across most OmniLLMs. For instance, MiniCPM2.6-o improves from 0.362 to 0.442 in accuracy on \emph{Caption Matching},
though it still lags behind the text-only setting. Overall, \dataset presents a new challenge and contributes to a comprehensive evaluation of MLLMs. It exposes their weakness in visual-centric humor understanding, stresses the need for enhancing their visual reasoning capabilities, and highlights the promise of integrating additional modalities like sound for video understanding.

\begin{figure*}[tp]
    \centering
    \includegraphics[width=\linewidth]{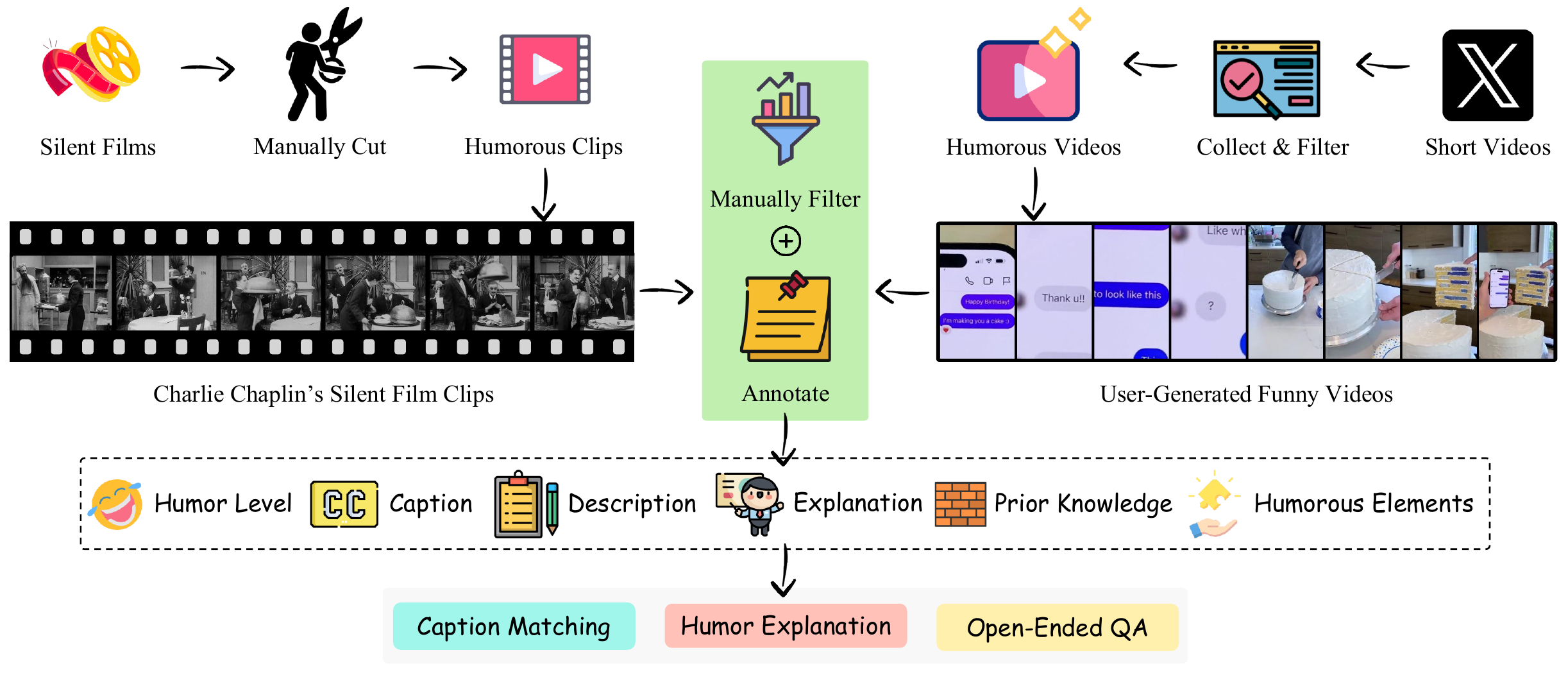}
    \caption{\label{fig:datasetPipeline} Data Curation Pipeline. To collect visual-centric humorous videos, the pipeline consists of two main stages: (a) \textit{Humorous video collection}, where annotators identify timestamps of self-contained humorous clips for silent films and verify humor presence in short videos (see Section~\ref{sec:data-collection}). (b) \textit{Filtering and annotation}, where only visual-dominant humor is retained and annotated (see Section~\ref{sec:data-annotation}). The annotation is further used for task construction (see Section~\ref{sec:tasks}).
    }
    \vspace{-1em}
\end{figure*}

\section{Visual-Centric Video Humor Curation}\label{sec:method}


\subsection{Humor Video Sources}\label{sec:data-collection}
Our goal is to collect humorous videos that are visual-centric and illustrate diverse humor. 
A straightforward approach is to collect humorous clips from silent comedies that are entirely devoid of speech. Though silent films may contain recorded music, sound effects, and few captions, which may contribute to the expression of humor, the humor primarily arises from the visual modality.
A major issue with silent film clips is that they have rather narrow themes and employ limited storytelling techniques. To enhance the diversity of humor in our dataset, we further incorporate user-generated short funny videos from the Internet. Specifically, we select videos primarily from an X account (@humansnocontext) that frequently shares humorous clips with minimal reliance on speech or text-based context.\footnote{Though most videos are sourced from a single X account, the content is sufficiently diverse (see Figure~\ref{fig:dataset}), as the account owner collects humorous videos from across the Internet.} We also use a subset of YouTube videos from~\citet{xie2024funqa} to diversify video sources.
Thus, our dataset comprises humorous videos from two different domains that complement each other (see Figure~\ref{fig:datasetPipeline}):

\begin{itemize}[leftmargin=*]
    \item \textbf{Charlie Chaplin’s Silent Films:} We reviewed Charlie Chaplin’s classic silent films from 1914 to 1938 and collected 729 funny clips. Each humor is ensured to be self-contained, without relying on additional video contexts. Figure \ref{fig:datasetPipeline} shows an example in this domain.
    \item \textbf{User-Generated Funny Videos:} We reviewed the X user \href{https://x.com/humansnocontext}{@humansnocontext}'s tweets posted between March 28, 2023 and October 12, 2024 and collected 18080 short funny videos. To ensure diverse video sources, we collected 1769 surprising YouTube videos from~\citet{xie2024funqa}.

\end{itemize}

\subsection{Preprocessing and Filtering}\label{sec:data-filter}
We preprocess and filter the initially collected videos according to duration, appropriateness, and speech reliance, sequentially.
(1) \textbf{Duration:} We retain videos ranging from 5 to 60 seconds long. Short clips under 5 seconds generally fail to convey meaningful humor, while clips exceeding 1 minute often rely on dialogue. For silent films, we segment long scenes to isolate individual humorous moments, ensuring that each segment captures the full humor, without becoming too long for generation tasks.
(2) \textbf{Appropriateness:} To ensure that the contents of our videos are appropriate, we adhered to the safety objectives outlined in 
\citet{thoppilan2022lamda} and excluded videos that violated the established criteria (see details in Appendix \ref{app:data-filter}).
(3) \textbf{Speech Reliance:} We minimize reliance on speech. Since there is little to no speech in Charlie Chaplin's silent films,
we primarily focused on user-generated funny videos and employed both manual and automatic approaches to filter out speech-heavy videos (see details in Appendix~\ref{app:data-filter}).

\begin{figure*}[tp]
    \centering
    \includegraphics[width=1.0\linewidth]{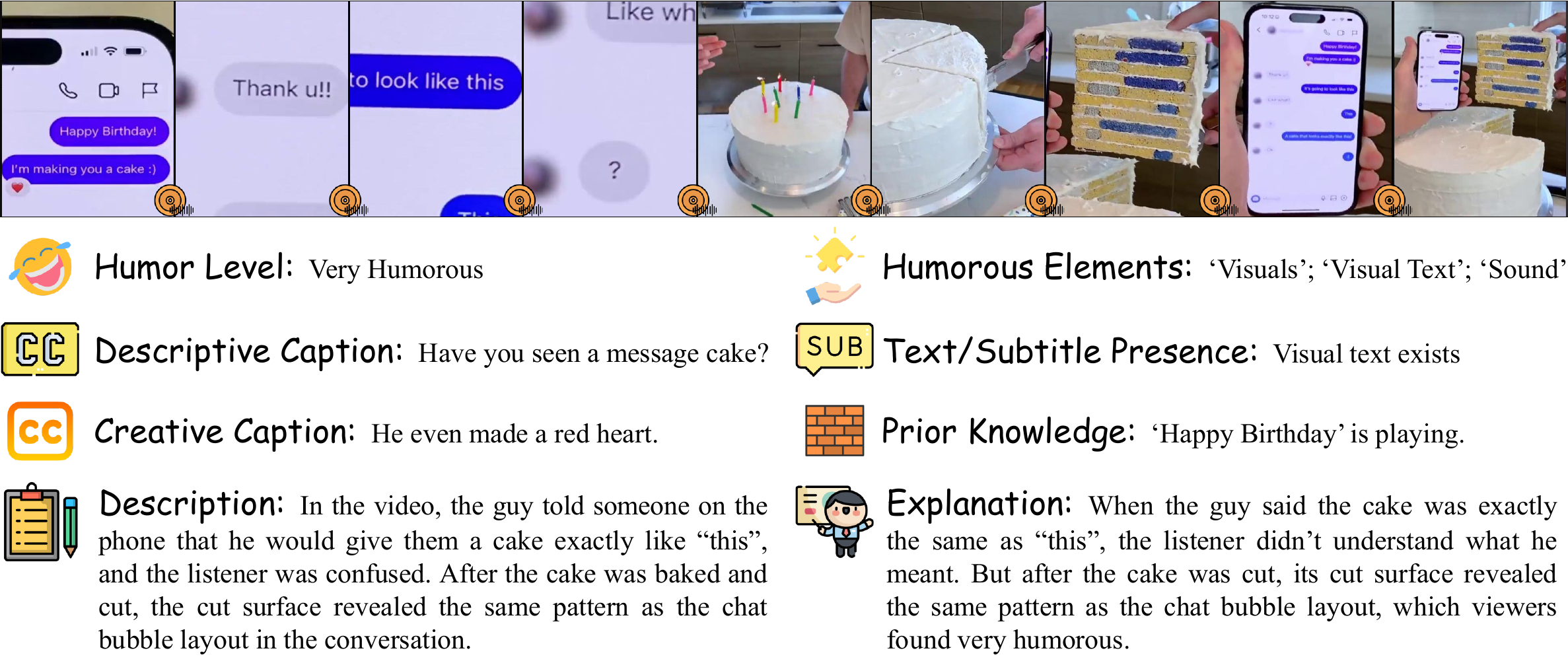}
    \caption{    \label{fig:mainfigure}Example annotation of a short video that conveys humor through visuals, visual text, and background sound. Knowing the Happy Birthday melody makes the video merrier (see Section~\ref{sec:data-annotation}).
   }
    \vspace{-1em}
\end{figure*}
\subsection{Annotation}\label{sec:data-annotation}
We recruited eight annotators based on the following criteria: (1) sufficient English proficiency to understand video content, (2) broad cultural knowledge to interpret humor arising from various contexts, and (3) strong observational skills assessed through a qualification test (see Appendix~\ref{app:data-annotate}). 
To ensure consistency, we provided detailed guidelines for each annotation task and created a reference manual for on-demand use. Each video underwent three rounds of annotation to guarantee correctness and thoroughness. We conducted the following primary annotation tasks (see Figure~\ref{fig:mainfigure} for an example annotation):
\begin{itemize}[leftmargin=*]
    \item \textbf{Humor Evaluation:}
    Annotators independently evaluated whether the video was humorous. 
    \item \textbf{Captioning:} Each annotator was asked to write two types of captions for each video, without seeing existing annotations, including captions and descriptions, from other annotators, thus ensuring an independent and unbiased judgment.
    \begin{itemize}[leftmargin=*]
    \item \emph{Descriptive Captions} directly describe or highlight the humor present in the video content from the original publisher’s perspective.
    \item \emph{Creative Captions} extend beyond the video’s original humor by adding imaginative or novel elements (see the visual caption in Figure~\ref{fig:teaser-b}).
    \end{itemize}
    The dual-caption annotation supports a comprehensive assessment of humor in video from both comprehension and generation perspectives. 
    \item \textbf{Video Description:} Annotators were instructed to describe the events in each video, including all the details necessary for understanding the humor, without making inferences, focusing only on observable objects, actions, and expressions. After the first annotator completes the video description, subsequent annotators review and refine the current descriptions for correctness and completeness.
    \item \textbf{Video Labeling:} Annotators labeled the key humor sources (e.g., human actions, objects, visual effects, or sound cues) in each video and noted whether any visual text was present. If an element appeared, but did not contribute to humor, it was not selected.
    \item \textbf{Background Knowledge:} \textcolor{black}{Annotators determine whether understanding the humor in a video requires background knowledge, which refers to external contextual information that cannot be directly inferred from the videos but is necessary or helpful for understanding humor.}
    \item \textbf{Humor Explanation:} Three annotators sequentially create and refine humor explanations by adding missing details, guaranteeing comprehensive coverage of the labeled humor sources through an iterative refinement process. In our annotation manual (see Figure~\ref{fig:annotation_manual} in Appendix) for humor explanation, we require that annotators include all the humorous elements they can find, as thoroughly as possible, and explain why these elements make people feel humorous.
\end{itemize}

\subsection{Data Analysis}\label{sec:data-analysis}
\begin{figure*}[t!]
    \centering
    \begin{subfigure}[b]{0.374\linewidth}
        \includegraphics[width=\linewidth]{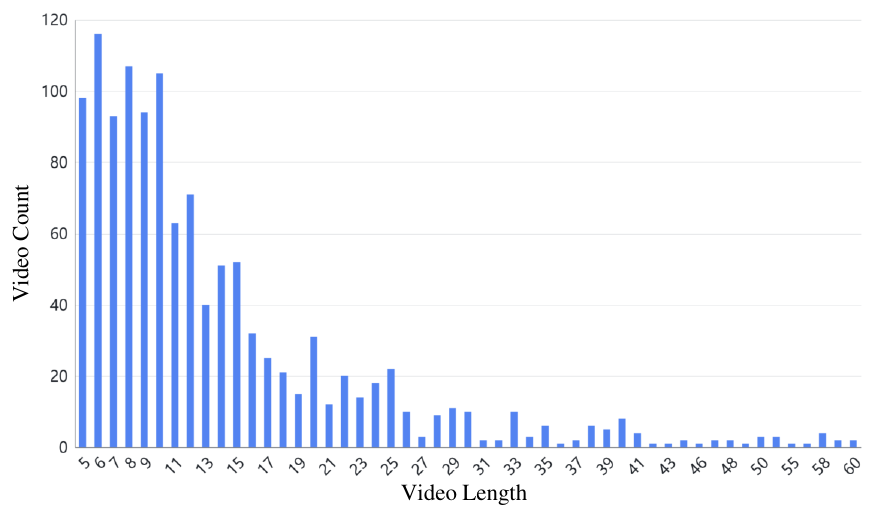}
        \caption{}
        \label{fig:data-duration}
    \end{subfigure}
    \hfill
    \begin{subfigure}[b]{0.181\linewidth}
        \includegraphics[width=\linewidth]{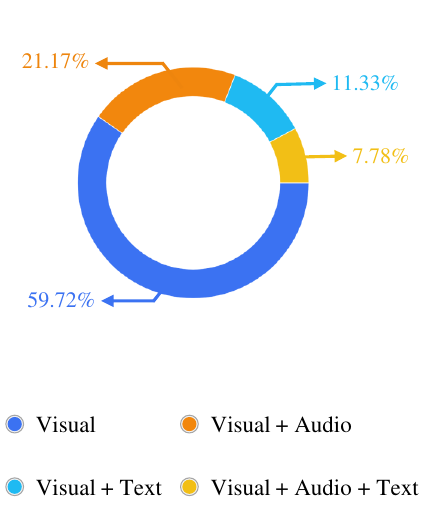}
        \caption{}
        \label{fig:data-modailty}
    \end{subfigure}
    \hfill
    \begin{subfigure}[b]{0.197\linewidth}
        \includegraphics[width=\linewidth]{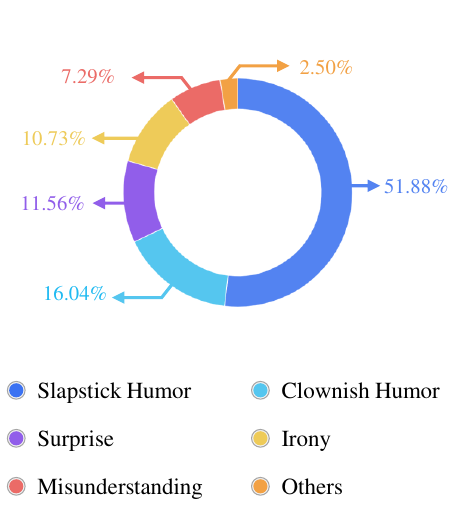}
        \caption{}
        \label{fig:data-type}
    \end{subfigure}
    \hfill
    \begin{subfigure}[b]{0.197\linewidth}
        \includegraphics[width=\linewidth]{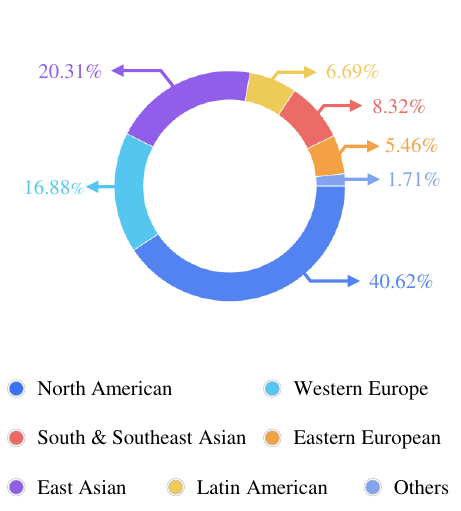}
        \caption{}
        \label{fig:data-culture}
    \end{subfigure}
    \caption{Data statistics: (a) Distribution of video lengths; (b) Distribution of visual-centric groups; (c) Distribution of humor types, where \emph{Others} contains Parody, Satire, Miscellaneous, etc; (d) Distribution of cultural backgrounds, where \emph{Others} covers Middle Eastern and North African, etc.}
    \label{fig:dataset}
\end{figure*}




\paragraph{Duration.} All videos in our final dataset are restricted to a duration of 5–60 seconds, with the majority concentrated within 30 seconds (see Figure~\ref{fig:data-duration}). This design ensures that humor is self-contained, sufficiently nuanced, and compatible with the context length limits of most MLLMs.

\paragraph{Diversity.} To show \dataset contains a variety of humor, we follow~\citet{buijzen2004developing} to categorize videos into five categories by humor type (see Figure \ref{fig:data-type}), and follow \citet{house2004culture} and~\citet{ronen2013mapping} to categorize videos into six categories by cultural background (see Figure \ref{fig:data-culture}). The illustrations demonstrate that \dataset covers a wide range of humor types and cultural backgrounds, thus it is sufficiently diverse.


\paragraph{Visual-centric.} After all filtering processes, we were left with 1218 videos, including 267 silent humor clips from Charlie Chaplin and 951 user-generated short funny videos from the Internet. The total duration of the videos is 4.7h, and the average duration is around 14s. All of them rely on the visual modality to express humor. We identify two key modalities that dominate the delivery of humor: visuals and audio. 
Apart from 722 videos (59\%) conveying humor primarily via pure visual cues (denoted by `Visual'), 137 videos (11\%) contain additional linguistic cues in visual form---such as embedded captions and subtitles (denoted by `Visual+Text')---that extend humor, 256 video humor (21\%) is enhanced by additional sound that covers non-speech auditory elements, such as background music and sound effects(denoted by `Visual+Audio'), and 94 videos (8\%) convey humor through visuals, sound, and visual text (denoted by `Visual+Audio+Text'). The video distribution over the four groups is illustrated in Figure~\ref{fig:data-modailty}.

\paragraph{Consensus Evaluation.} For annotations like humor explanation and video description, the second and third annotators reviewed and modified previous annotations to ensure consistency. We employ Krippendorff’s alpha \citep{krippendorff2011computing} to assess the annotators' consensus on the humor evaluation, using `Low,' `Medium,' and `High' to indicate the strength of the consensus. In \dataset, more than 90\% of data demonstrated a `High' consensus, while only 0.3\% showed a `Low' consensus.
\section{\model: A Visual-Centric Video Humor Understanding Benchmark}\label{sec:tasks}


\subsection{Evaluation Tasks}

To comprehensively evaluate the capability of MLLMs in humor understanding, we propose three tasks that reflect different aspects of humor reasoning: \text{Caption Matching}, \text{Humor Explanation}, and \text{Open-ended QA}:

\begin{itemize}[leftmargin=*]
\item\textbf{Caption Matching.}
In this discriminative task, models must correctly associate videos with their corresponding captions.
Unlike ordinary caption matching tasks, our design challenges MLLMs to go beyond surface-level matching and assess their ability to understand video humor that is pronounced by \emph{creative captions} from a generation perspective. For each video with a creative caption, we randomly sample four \emph{descriptive captions} from other videos as the distractors.

\item\textbf{Humor Explanation.} 
In this generative task, models must identify humor points within each video, provide coherent explanations, and reference relevant visual or auditory cues. 

\item\textbf{Open-ended QA.} To further assess the fundamental understanding of video content, we generate a set of open-ended question-answer pairs for each video (see details in Appendix~\ref{app:generate_qapair}). These questions—automatically generated by GPT-4o and manually verified—encompass temporal, descriptive, and causal aspects \citep{xiao2021next}.\footnote{There are 81, 742, and 395 QA pairs for temporal, descriptive, and causal questions, respectively.} This extends the benchmark beyond humor-specific reasoning, providing a broader assessment of video reasoning skills.
\end{itemize}

\subsection{Evaluation Methods}

We employ different evaluation strategies depending on the task type:

\begin{itemize}[leftmargin=*]
\item \textbf{Accuracy.} For \emph{Caption Matching}, we measure accuracy to determine whether the model correctly identifies the most appropriate response.

\item \textbf{Quality of Open-ended Responses.} For \emph{Humor Explanation} and \emph{Open-ended QA}, we adopt both automatic and human evaluation approaches:
\begin{itemize}[leftmargin=*]
    \item \emph{Semantic Similarity.} We compute similarity scores between model-generated answers and human-provided answers using BERTScore \citep{Zhang*2020BERTScore:}, which captures fine-grained semantic similarity beyond simple word overlap. In addition, we employ SentBERT~\citep{reimers2019sentence} to assess sentence-level semantic coherence, as well as METEOR~\citep{banerjee-lavie-2005-meteor}, which provides a more nuanced assessment of semantic adequacy and fluency. 
    
    \item \textcolor{black}{AutoDQ \citep{wang2024tarsier}: This method evaluates the presence of humor-related events in the generated explanations. AutoDQ extracts key events from the model’s output and compares them to ground truth (GT) annotations using entailment analysis. It provides three metrics: recall, precision, and F1 score (see Appendix~\ref{app:eval-method} for details).}  \textcolor{black}{Unless otherwise specified, we report F1 scores.}
    
    \item \emph{Human Evaluation.} 
    We randomly sample a subset of model-generated explanations and compare them with ground truth. The evaluators rate the explanations based on accuracy and logicality, providing insight into the gap between human and MLLMs' explanations. We present and discuss the results in Table~\ref{tab:human_preference} in  Appendix~\ref{app:more-results} due to limited space.

\end{itemize}
\end{itemize}

\section{Experiments}\label{sec:experiments}

\subsection{Experimental Setup}
\label{sec:exp-setup}

\paragraph{MLLMs.} We consider both proprietary MLLMs, such as Gemini-2.5-Flash~\citep{gemini25} and GPT-4o~\citep{hurst2024gpt}, as well as public MLLMs like Qwen2.5-VL (72B)~\citep{qwen25vl}, and Intern3.5-VL (8B)~\citep{wang2025internvl3_5}. OmniLLMs such as Video-SALMONN-2 (7B)~\citep{video-salmon2}, MiniCPM2.6-o (8B)~\citep{yao2024minicpm}, and Qwen2.5-Omni (7B)~\citep{qwen2.5-omni}, which can process audio, are also included (an overview of all evaluated MLLMs is presented in Table~\ref{tab:model_list}).

\paragraph{Evaluation Settings.} To understand the roles of different modalities in video humor understanding, we consider the following three settings: Text-Only, Video-Only, and Video+Audio, which means models are tested with text, video (w/ audio), and video-audio inputs, respectively. 

\begin{itemize}[leftmargin=*]

\item \emph{Text-Only.} In this setting, models receive detailed human-written video descriptions; no visual or audio information is available to the models. Thus, it evaluates the language reasoning ability of MLLMs in isolation.

\item \emph{Video-Only.} Models are provided with only raw video frames (w/o audio). This setting assesses their intrinsic visual comprehension capabilities. Depending on the presence of visual text, we further divide results into two groups: `w/ visual text' and `w/o visual text'. 

\item \emph{Video+Audio.} Models receive both video frames and audio signals, allowing us to examine whether the inclusion of auditory information improves humor understanding. Depending on the contribution of audio to humor, we further divide results into two groups: `w/ humor audio' and `w/o humor video'.
    
\end{itemize}

\begin{table}[tb]
\centering
\caption{
Model performance on three tasks.
}
\label{tab:main-result}
\resizebox{\linewidth}{!}{%
\begin{tabular}{ccclccc}
\hline
\multicolumn{1}{c|}{}                        & \multicolumn{3}{c|}{Explanation}                & \multicolumn{1}{c|}{Matching} & \multicolumn{2}{c}{Open-ended QA} \\ \cline{2-7} 
\multicolumn{1}{c|}{\multirow{-2}{*}{MLLMs}} & SentBERT & METEOR & \multicolumn{1}{l|}{AutoDQ} & \multicolumn{1}{c|}{Accuracy} & SentBERT         & METEOR         \\ \hline
\multicolumn{7}{c}{\cellcolor[HTML]{EFEFEF}\textit{Text-Only}}                                                                                                              \\ \hline
\multicolumn{1}{c|}{Gemini-2.5-Flash}        & 0.547    & 0.249  & \multicolumn{1}{l|}{0.342}  & \multicolumn{1}{c|}{0.615}    & 0.728            & 0.642          \\
\multicolumn{1}{c|}{Video-SALMONN-2}         & 0.571    & 0.246  & \multicolumn{1}{l|}{0.317}  & \multicolumn{1}{c|}{0.359}    & 0.595            & 0.435          \\
\multicolumn{1}{c|}{MiniCPM2.6-o}            & 0.546    & 0.236  & \multicolumn{1}{l|}{0.325}  & \multicolumn{1}{c|}{0.531}    & 0.562            & 0.454          \\
\multicolumn{1}{c|}{Qwen2.5-Omni}          & 0.536    & 0.233  & \multicolumn{1}{l|}{0.316}  & \multicolumn{1}{c|}{0.656}    & 0.719            & 0.546          \\
\multicolumn{1}{c|}{Qwen2.5-VL}         & 0.543    & 0.250  & \multicolumn{1}{l|}{0.342}  & \multicolumn{1}{c|}{0.726}    & 0.760            & 0.598          \\
\multicolumn{1}{c|}{Intern3.5-VL}            & 0.556    & 0.256  & \multicolumn{1}{l|}{0.348}  & \multicolumn{1}{c|}{0.643}    & 0.701            & 0.593          \\
\multicolumn{1}{c|}{GPT-4o}                  & 0.560    & 0.255  & \multicolumn{1}{l|}{0.374}  & \multicolumn{1}{c|}{0.762}    & 0.690            & 0.645          \\ \hline
\multicolumn{7}{c}{\cellcolor[HTML]{EFEFEF}\textit{Video-Only}}                                                                                                             \\ \hline
\multicolumn{1}{c|}{Gemini-2.5-Flash}        & 0.459    & 0.199  & \multicolumn{1}{l|}{0.175}  & \multicolumn{1}{c|}{0.580}    & 0.424            & 0.270          \\
\multicolumn{1}{c|}{video-SALMONN-2}         & 0.269    & 0.150  & \multicolumn{1}{l|}{0.052}  & \multicolumn{1}{c|}{0.243}    & 0.317            & 0.169          \\
\multicolumn{1}{c|}{MiniCPM2.6-o}            & 0.381    & 0.165  & \multicolumn{1}{l|}{0.112}  & \multicolumn{1}{c|}{0.362}    & 0.369            & 0.186          \\
\multicolumn{1}{c|}{Qwen2.5-Omni}           & 0.384    & 0.159  & \multicolumn{1}{l|}{0.144}  & \multicolumn{1}{c|}{0.553}    & 0.382            & 0.121          \\
\multicolumn{1}{c|}{Qwen2.5-VL}         & 0.441    & 0.187  & \multicolumn{1}{l|}{0.150}  & \multicolumn{1}{c|}{0.666}    & 0.445            & 0.202          \\
\multicolumn{1}{c|}{Intern3.5-VL}            & 0.422    & 0.180  & \multicolumn{1}{l|}{0.125}  & \multicolumn{1}{c|}{0.609}    & 0.385            & 0.235          \\
\multicolumn{1}{c|}{GPT-4o}                  & 0.455    & 0.192  & \multicolumn{1}{l|}{0.206}  & \multicolumn{1}{c|}{0.646}    & 0.411            & 0.286          \\ \hline
\multicolumn{7}{c}{\cellcolor[HTML]{EFEFEF}\textit{Video+Audio}}                                                                                                            \\ \hline
\multicolumn{1}{c|}{Gemini-2.5-Flash}        & 0.460    & 0.199  & \multicolumn{1}{l|}{0.173}  & \multicolumn{1}{c|}{0.581}    & 0.416            & 0.268          \\
\multicolumn{1}{c|}{video-SALMONN-2}         & 0.281    & 0.170  & \multicolumn{1}{l|}{0.066}  & \multicolumn{1}{c|}{0.240}    & 0.323            & 0.185          \\
\multicolumn{1}{c|}{MiniCPM2.6-o}            & 0.408    & 0.173  & \multicolumn{1}{l|}{0.120}  & \multicolumn{1}{c|}{0.442}    & 0.380            & 0.278          \\
\multicolumn{1}{c|}{Qwen2.5-Omni}          & 0.428    & 0.174  & \multicolumn{1}{l|}{0.125}  & \multicolumn{1}{c|}{0.617}    & 0.424            & 0.168          \\ \hline
\end{tabular}
}
\end{table}

\subsection{Main Results}

Based on the results in \Cref{tab:main-result}, we analyze the humor competence of MLLMs along three dimensions: video humor discovery, understanding, and subtle humor inference. Our results reveal several shortcomings of MLLMs: they (i) struggle to identify humorous elements when explicit cues are absent, (ii) inadequately fuse information across modalities for understanding, and (iii) show limited capacity for inferring subtle humor.

\paragraph{Limited ability in humor discovery.}
Across settings, MLLMs tend to perform better on open-ended QA than on humor explanation. This performance disparity reveals that they are limited in perceiving humor. For example, in the Text-Only setting, Qwen2.5-VL, whose SentBERT score drops from 0.760 in QA to 0.543 in humor explanation. These findings suggest that MLLMs are more successful when the question itself provides explicit cues that direct attention to a specific humorous element in the scene. By contrast, the humor explanation task, which requires models to independently identify and articulate the source of humor without such guidance, poses a greater challenge. This indicates that while MLLMs are often able to reason about humor once it is highlighted for them, they struggle with the more cognitively-demanding task of discovering humor directly from contextual cues.

\paragraph{Heavy reliance on linguistic cues for humor understanding.}
Comparing text-based and video-based evaluations, we observe marked differences across all three tasks, where the Text-Only setting yields substantially higher scores than the video-based settings, implying that current MLLMs are heavily dependent on linguistic cues for humor understanding. For example, on open-ended QA, Qwen2.5-VL achieves a SentBERT score of 0.760 with text input, but it plummets to 0.445 when presented with raw video (w/o audio). While the addition of audio provides a marginal but consistent performance boost, this gain is minimal compared to the contribution of text. This wide performance gap suggests that MLLMs' cross-modal fusion capabilities are still underdeveloped, leading them to rely on linguistic cues rather than effectively integrating visual and auditory signals. 

\paragraph{Incapability for subtle humor inference.} 
The caption matching task goes beyond surface-level linking between literal descriptions and videos; instead, it requires models to find the \emph{creative caption} that enhances or extends humor in the video. We find that most models exhibit limited performance (e.g., below 0.8), suggesting their incompetence for subtle humor inference. The difficulty is magnified when models process raw video. For example, video-SALMONN-2's accuracy falls from 0.359 in the Text-Only setting to 0.240 in the Video+Audio condition. This pronounced struggle to connect creative text to original visual humor context reveals a critical weakness in the models' capacity for the implicit cross-modal reasoning that is fundamental to comprehending sophisticated humor.

\subsection{Further Analysis}

\begin{table}[t!]
\centering
\caption{
The impact of visual text on video humor understanding in the Video+Audio setting.}
\label{tab:visual_sound}
\resizebox{\linewidth}{!}{%
\begin{tabular}{@{}ccccccc@{}}
\toprule
\multicolumn{1}{c|}{}                         & \multicolumn{3}{c|}{Sound contributing to humor}                       & \multicolumn{3}{c}{Sound not contributing to humor}                \\ \cmidrule(l){2-7} 
\multicolumn{1}{c|}{}                         & \multicolumn{1}{c|}{Matching} & \multicolumn{2}{c|}{Open-ended QA}     & \multicolumn{1}{c|}{Matching}  & \multicolumn{2}{c}{Open-ended QA} \\ \cmidrule(l){2-7} 
\multicolumn{1}{c|}{\multirow{-3}{*}{Models}} & \multicolumn{1}{c|}{Accuracy} & SentBERT & \multicolumn{1}{c|}{METEOR} & \multicolumn{1}{c|}{Accuracy}  & SentBERT          & METEOR        \\ \midrule
\multicolumn{7}{c}{\cellcolor[HTML]{EFEFEF}\textit{w/ visual text}}                                                                                                                                  \\ \midrule
\multicolumn{1}{c|}{Gemini-2.5-Flash}         & \multicolumn{1}{c|}{0.621}    & 0.440    & \multicolumn{1}{c|}{0.303}  & \multicolumn{1}{c|}{0.715}     & 0.437             & 0.288         \\
\multicolumn{1}{c|}{video-SALMONN-2}          & \multicolumn{1}{c|}{0.200}    & 0.336    & \multicolumn{1}{c|}{0.202}  & \multicolumn{1}{c|}{0.255}     & 0.332             & 0.193         \\
\multicolumn{1}{c|}{MiniCPM2.6-o}             & \multicolumn{1}{c|}{0.453}    & 0.466    & \multicolumn{1}{c|}{0.357}  & \multicolumn{1}{c|}{0.511}     & 0.430             & 0.320         \\
\multicolumn{1}{c|}{Qwen2.5-Omni}            & \multicolumn{1}{c|}{0.716}    & 0.442    & \multicolumn{1}{c|}{0.174}  & \multicolumn{1}{c|}{0.686}     & 0.434             & 0.171         \\ \midrule
\multicolumn{7}{c}{\cellcolor[HTML]{EFEFEF}\textit{w/o visual text}}                                                                                                                                 \\ \midrule
\multicolumn{1}{c|}{Gemini-2.5-Flash}         & \multicolumn{1}{c|}{0.523}    & 0.396    & \multicolumn{1}{c|}{0.257}  & \multicolumn{1}{c|}{0.569}     & 0.416             & 0.265         \\
\multicolumn{1}{c|}{video-SALMONN-2}          & \multicolumn{1}{c|}{0.235}    & 0.301    & \multicolumn{1}{c|}{0.174}  & \multicolumn{1}{c|}{0.245}     & 0.327             & 0.185         \\
\multicolumn{1}{c|}{MiniCPM2.6-o}             & \multicolumn{1}{c|}{0.378}    & 0.369    & \multicolumn{1}{c|}{0.280}  & \multicolumn{1}{c|}{0.449}     & 0.338             & 0.215         \\
\multicolumn{1}{c|}{Qwen2.5-Omni}            & \multicolumn{1}{c|}{0.547}    & 0.432    & \multicolumn{1}{c|}{0.151}  & \multicolumn{1}{c|}{0.612}     & 0.415             & 0.171         \\ \bottomrule
\end{tabular}
}
\vspace{0em}
\end{table}

To conduct a deeper analysis of model results, we further divide our experimental results based on previously annotated humor modalities and background knowledge essential for delivering humor in video, to analyze how different types of humor affect models' explanatory capability. We also presented supplementary quantitative results in Appendix~\ref{app:more-results}, such as human preference comparison of humor explanations across four model categories. (see Table~\ref{tab:human_preference}).

\paragraph{Visual text adds value regardless of the audio's comedic contribution.} As shown in Table~\ref{tab:visual_sound}, MLLMs generally perform better on videos containing visual text than on those without linguistic cues in the Video+Audio setting, except for video-SALMONN-2 on the caption matching task when sound contributes to humor. For example, Gemini-2.5-Flash achieves SentBERT and METEOR scores of 0.440 and 0.303 for open-ended QA with visual text, compared to 0.396 and 0.257 without it. When sound does not contribute to humor, the benefit of visual text remains evident: Gemini-2.5-Flash improves from 0.416 to 0.437 in open-ended QA SentBERT and from 0.569 to 0.715 in matching accuracy with visual text. These results indicate that \emph{visual text serves as a useful complementary cue for humor understanding}, particularly when audio signals provide limited informative content.

\paragraph{Knowledge-based cues facilitate humor understanding.} We identified 374 videos that require contextual background knowledge and evaluated MLLMs under two settings: with and without the explicit provision of such knowledge. As shown in Table~\ref{tab:with_background}, MLLMs consistently achieve higher performance when background knowledge is provided under the Video+Audio setting. For instance, Qwen2.5-Omni attains a SentBERT and AutoDQ scores of 0.512 and 0.176 on the humor explanation with background knowledge, compared to 0.459 and 0.127 without. These findings suggest that while MLLMs implicitly encode multiple cultural contexts, their \emph{comprehension of humor is significantly enhanced by the explicit provision of background knowledge}, underscoring the central role of linguistic and knowledge-based cues in complex video humor understanding tasks.

\begin{table}[]
\caption{
The impact of background knowledge on video humor understanding in the Video+Audio setting.
}
\label{tab:with_background}
\resizebox{\linewidth}{!}{%
\begin{tabular}{@{}ccclccc@{}}
\toprule
\multicolumn{1}{c|}{}                        & \multicolumn{3}{c|}{Explanation}                & \multicolumn{1}{c|}{Matching} & \multicolumn{2}{c}{Open-ended QA} \\ \cmidrule(l){2-7} 
\multicolumn{1}{c|}{\multirow{-2}{*}{MLLMs}} & SentBERT & METEOR & \multicolumn{1}{l|}{AutoDQ} & \multicolumn{1}{c|}{Accuracy} & SentBERT         & METEOR         \\ \midrule
\multicolumn{7}{c}{\cellcolor[HTML]{EFEFEF}\textit{w/ Background Knowledge}}                                                                                             \\ \midrule
\multicolumn{1}{c|}{video-SALMONN-2}         & 0.467    & 0.173  & \multicolumn{1}{l|}{0.114}  & \multicolumn{1}{c|}{0.324}    & 0.396            & 0.198          \\
\multicolumn{1}{c|}{MiniCPM2.6-o}           & 0.515    & 0.203  & \multicolumn{1}{l|}{0.193}  & \multicolumn{1}{c|}{0.447}    & 0.427            & 0.204          \\
\multicolumn{1}{c|}{Qwen2.5-Omni}           & 0.512    & 0.199  & \multicolumn{1}{l|}{0.176}  & \multicolumn{1}{c|}{0.663}    & 0.493            & 0.219          \\ \midrule
\multicolumn{7}{c}{\cellcolor[HTML]{EFEFEF}\textit{w/o Background Knowledge}}                                                                                               \\ \midrule
\multicolumn{1}{c|}{video-SALMONN-2}         & 0.285    & 0.177  & \multicolumn{1}{l|}{0.025}  & \multicolumn{1}{c|}{0.252}    & 0.301            & 0.174          \\
\multicolumn{1}{c|}{MiniCPM2.6-o}           & 0.440    & 0.180  & \multicolumn{1}{l|}{0.115}  & \multicolumn{1}{c|}{0.417}    & 0.358            & 0.259          \\
\multicolumn{1}{c|}{Qwen2.5-Omni}           & 0.459    & 0.181  & \multicolumn{1}{l|}{0.127}  & \multicolumn{1}{c|}{0.615}    & 0.441            & 0.157          \\ \bottomrule
\end{tabular}
}
\vspace{-1.5em}
\end{table}

\paragraph{MLLMs have greater difficulty in comprehending humor in historically distant videos.} We analyze the performance of MLLMs under the Video-Only setting across two subsets from distinct eras: last-century Charlie Chaplin's silent films (CCSF) and contemporary user-generated funny videos (UGFV). As shown in Table~\ref{tab:ccsf-vs-ugfv}, MLLMs consistently achieve higher scores on UGFV across all evaluation metrics. For example, Gemini-2.5-Flash attains a SentBERT of 0.469 for humor explanation and 0.434 for open-ended QA on UGFV videos, compared to 0.422 and 0.386, respectively, on CCSF videos. These findings suggest that MLLMs face greater difficulty in comprehending humor in historically distant videos, \emph{highlighting the sensitivity of humor understanding to the temporal and cultural context of videos.}

\begin{table}[t]
\caption{The impact of video era on video humor understanding in the Video-Only setting.}
\label{tab:ccsf-vs-ugfv}
\resizebox{1.0\linewidth}{!}{%
\begin{tabular}{@{}ccclccc@{}}
\toprule
\multicolumn{1}{c|}{\multirow{2}{*}{MLLMs}} & \multicolumn{3}{c|}{Explanation}                & \multicolumn{1}{c|}{Matching} & \multicolumn{2}{c}{Open-ended QA} \\ \cmidrule(l){2-7}  
\multicolumn{1}{c|}{}                       & SentBERT & METEOR & \multicolumn{1}{l|}{AutoDQ} & \multicolumn{1}{c|}{Accuracy} & SentBERT         & METEOR         \\ \midrule
\multicolumn{7}{c}{\cellcolor[HTML]{EFEFEF}\textit{\textbf{Last-Century} Charlie Chaplin's Silent Films}}                                                                                                   \\ \midrule
\multicolumn{1}{c|}{Gemini-2.5-Flash}       & 0.422    & 0.188  & \multicolumn{1}{l|}{0.130}  & \multicolumn{1}{c|}{0.562}    & 0.386            & 0.221          \\
\multicolumn{1}{c|}{video-SALMONN-2}        & 0.281    & 0.146  & \multicolumn{1}{l|}{0.012}  & \multicolumn{1}{c|}{0.165}    & 0.296            & 0.154          \\
\multicolumn{1}{c|}{MiniCPM2.6-o}           & 0.343    & 0.150  & \multicolumn{1}{l|}{0.097}  & \multicolumn{1}{c|}{0.307}    & 0.314            & 0.128          \\
\multicolumn{1}{c|}{Qwen2.5-Omni}           & 0.339    & 0.144  & \multicolumn{1}{l|}{0.096}  & \multicolumn{1}{c|}{0.494}    & 0.337            & 0.119          \\ \midrule
\multicolumn{7}{c}{\cellcolor[HTML]{EFEFEF}\textit{\textbf{Contemporary} User-Generated Funny Video}}                                                                                                       \\ \midrule
\multicolumn{1}{c|}{Gemini-2.5-Flash}       & 0.469    & 0.202  & \multicolumn{1}{l|}{0.194}  & \multicolumn{1}{c|}{0.586}    & 0.434            & 0.283          \\
\multicolumn{1}{c|}{video-SALMONN-2}        & 0.265    & 0.151  & \multicolumn{1}{l|}{0.061}  & \multicolumn{1}{c|}{0.265}    & 0.322            & 0.174          \\
\multicolumn{1}{c|}{MiniCPM2.6-o}           & 0.392    & 0.170  & \multicolumn{1}{l|}{0.118}  & \multicolumn{1}{c|}{0.378}    & 0.384            & 0.203          \\
\multicolumn{1}{c|}{Qwen2.5-Omni}           & 0.397    & 0.164  & \multicolumn{1}{l|}{0.166}  & \multicolumn{1}{c|}{0.570}    & 0.395            & 0.121          \\ \bottomrule
\end{tabular}
}
\vspace{-1.2em}
\end{table}

\begin{table}[ht]
\centering
\caption{Comparison between MLLMs and their base LLMs under the Text-Only setting.}
\label{tab:desc-only-base-vs-ml}
\resizebox{\linewidth}{!}{%
\begin{tabular}{@{}l|ccc@{}}
\toprule
\multirow{2}{*}{Models} & \multicolumn{3}{c}{Open-ended QA}                 \\ \cmidrule(l){2-4} 
                            & SentBERT & METEOR & BERTScore \\ \midrule
Qwen2.5-VL              & 0.760    & 0.598  & 0.730    \\
Qwen2.5-72B                 & 0.692    & 0.624  & 0.690    \\
\toprule
Qwen2.5-Omni             & 0.719    & 0.540  & 0.687    \\
Qwen2.5-7B                  & 0.674    & 0.526  & 0.657    \\ \bottomrule
\end{tabular}
}
\vspace{-1.2em}
\end{table}
\paragraph{MLLMs vs. their base LLMs.} MLLMs are usually derived from a pre-trained base LLM by adding a visual encoder or multimodal modules. For instance, Qwen2.5-VL extends Qwen2.5-72B, and Qwen2.5-Omni extends Qwen2.5-7B (see Table~\ref{tab:desc-only-base-vs-ml}). In the Text-Only setup, Qwen2.5-Omni surpasses Qwen2.5-7B with a SentBERT score of 0.719 (vs. 0.674) and a BERTScore score of 0.687 (vs. 0.657) on open-ended QA task, suggesting that \emph{multimodal training can confer advantages even when only textual descriptions are available, possibly because the model has learned richer contextual associations during training}. Please refer to Table~\ref{tab:add_mllmvsllm} for more details on humor explanation and caption matching tasks.

\section{Related Work}\label{sec:related}


\paragraph{From LLMs to Video LLMs.} Large language models have demonstrated outstanding capabilities in many domains, including natural language processing, coding, math, and reasoning, ushering in new breakthroughs for video understanding technology. Video LLMs integrate visual encoders with LLMs, leading to a unified model to
reason across video and language in the same language space~\citep{wang2024qwen2, liu2024oryx, Lin2023VideoLLaVALU}.
Early video LLMs employ pre-trained image encoder and video encoder to encode only video frames ~\citep{zhang2023videollama, maaz2023videochatgpt, Li2023VideoChatCV, Lin2023VideoLLaVALU}.
Recent works augment video LLMs with an audio encoder to align visual, auditory, and textual modalities in the same language space. Moreover, the audio encoder is supposed to capture diverse environmental sound apart from human speech since sound has been shown to contain amounts of commonsense knowledge ~\citep{videollama2, qwen2.5-omni}.

\paragraph{Video LLMs.} 
Video LLMs have shown remarkable performance in many traditional video processing tasks such as video captioning~\citep{xu2016msr, agrawal2019nocaps, 10.1007/s11263-016-0965-7}, video question answering \citep{VQA,xiao2021next, yu2019activityqa, video-mme}, and grounding \citep{kazemzadeh-etal-2014-referitgame,wu2022grit}. However, most existing benchmarks primarily target general video understanding tasks, such as MVBench~\citep{mvbench}, Video-MME~\citep{video-mme}, PerceptionTest~\citep{perceptiontest}, MLVU~\citep{mlvu}, and LVBench~\citep{lvbench}, which mainly assess the recognition of basic visual cues across videos of varying lengths. Others are designed to evaluate specific video understanding capabilities, including temporal grounding \citep{charades-sta, qvhighlight, didemo, hawkeye}, video object detection \citep{vidor, vidvrd}, and video hallucination \citep{videohallucer, cmm}. But there remains a pressing need for benchmarks that evaluate higher-level cognitive abilities, such as social intelligence, to better measure the gap between human and MLLMs' performance. 

Our work narrows this gap. We expand the evaluation spectrum of video LLMs by introducing a challenging evaluation framework for humor understanding, formulating a humor generation task, and presenting the first comprehensive humor-centric evaluation and analysis.

\paragraph{Humor Video Understanding.} Humor understanding is a popular research topic of artificial intelligence and its roots in cognitive science~\citep{Hampes_2001,Hampes_2010}. While early works focus on verbal humor like jokes and sarcasm,~\citep{joke1,joke2,sarcasm}, the advent of LLMs has extended this scope to multimodal humor with image-language and video-language humor~\citep{hessel2022androids,alnajjar-etal-2022-laugh}. However, a parallel evaluation of video LLMs is still limited. 

While there have been several video-based humor datasets, the humor within them either is primarily dominated by spoken dialogue~\citep{kumar-etal-2022-become,hyun-etal-2024-smile} or is restricted to those that can only be understood when both visual and linguistic cues are present~\citep{ko2023can}, ignoring the fact that humans can appreciate humor solely from visuals.
HumorQA~\citep{xie2024funqa} is most similar to ours, but it is restricted to surprising videos only and has an average duration of 7 seconds; consequently, it is limited in coverage and temporal complexity and is insufficient for humor-centric video understanding evaluation. Moreover, it overlooked environmental sound that generally enhances humor. In contrast, we focus on humor understanding and cover more diverse video humor, with an average duration twice that of HumorQA.



Sound has been found informative of commonsense~\citep{zhao-etal-2022-connecting,Zellers_2022_CVPR}, and it has been shown that integrating textual, acoustic, and visual features significantly improves humor detection accuracy~\citep{chandrasekaran2016we, hasan-etal-2019-ur}. Since multimodal LLMs have recently been extended to natively support audio processing (aka. OmniLLMs), we propose and conduct a first evaluation of MLLMs on video humor understanding that involves environmental sound.

\section{Conclusion}

We have introduced \dataset, a video humor understanding benchmark from vision and sound. \dataset is designed to assess and diagnose the capability of MLLMs for video humor understanding. Each video is annotated with captions, descriptions, explanations, etc., supporting evaluation tasks such as caption matching, humor explanation, and open-ended QA. We evaluated a diverse range of MLLMs, spanning open-sourced and proprietary domains and covering specialized video LLMs and versatile OmniLLMs. Our findings reveal that current MLLMs heavily rely on linguistic cues for humor understanding, but are weak in deriving nuanced visual cues for understanding sophisticated video humor. Moreover, we empirically find that including environmental sound helps with humor understanding, highlighting the informativeness of sound and the promise of incorporating rich modalities for complex video reasoning tasks.

\label{sec:conclusion}

\section*{Limitations}

We acknowledge two limitations of this work: (1) \dataset contains humorous videos from diverse cultural backgrounds. Although culture-level categorization would be more fine-grained, assigning a single culture label to visual-centric humorous videos is often ambiguous and subjective, due to the prevalence of hybrid and globally shared visual humor. We therefore adopt region-level categorization, which is proposed by \citet{house2004culture} and~\citet{ronen2013mapping}, as a more stable, weakly supervised proxy for cultural background, while acknowledging that a geographical region may encompass multiple cultures. (2) Second, all data annotations were created manually in English. However, the ability of LLMs to understand humor may vary across languages, as different languages interpret and express the same visual content in distinct ways. This points to a promising direction for future research.

\section*{Ethical Considerations}
We firmly adhere to the ACL Code of Ethics in the performance of this work and the methods involved. We respect intellectual property and privacy rights and restrict the dataset to non-commercial academic use (see more details in Appendix~\ref{app:copyright}). Additionally, we carefully screened user-generated content to ensure the benchmark is safe for the research community and minimizes privacy risks for private individuals. Moreover, LLMs may produce offensive and incorrect statements, explicitly warn against the misuse of this dataset for generating malicious or mocking content, advocating instead for the development of safe and empathetic AI systems. This work is released with the intent for research purposes only.

\section*{Acknowledgements}
\textcolor{black}{Yanpeng Zhao acknowledges the support of the National Natural Science Foundation of China (12574467). Zilong Zheng is supported by the National Natural Science Foundation of China (62376031). We would like to thank Hengli Li for their helpful suggestions. We are also grateful to all annotators who contributed to the construction and verification of \model. Their help and efforts were essential to ensuring the quality and reliability of the benchmark.}

\paragraph{Use of LLMs.} In this work, we used LLMs as assistive tools in the following stages:
\begin{itemize}[leftmargin=*]
    \item \textbf{Dataset Construction.} We initially employed GPT-4o~\citep{hurst2024gpt} to assist in generating candidate QA pairs and humor categories from video content. All outputs were subsequently reviewed and revised by human annotators to ensure correctness and quality.
    \item \textbf{Code Assistance.} LLMs were used to help generate parts of the evaluation code, which were then verified and refined by the authors.
    \item \textbf{Writing Support.} ChatGPT was used to write and polish some sentences in Section~\ref{sec:experiments} for readability.
\end{itemize}


\bibliography{ACL2026_arxiv/sec/11_references}

\appendix

\clearpage


\section*{\centering Overview of Appendix}
This appendix provides supplementary materials that support the main paper. It includes (1) data curation and annotation procedures in Appendix~\ref{sec:crowdworking}, (2) extended descriptions of task definitions and evaluation settings in Appendix~\ref{app:add_exp_note}, (3) supplementary quantitative results in Appendix~\ref{app:more-results}, (4) qualitative case studies in Appendix~\ref{sec:case_study}, (5) and the prompts used for evaluating MLLMs in Appendix~\ref{sec:prompts}. These materials are intended to enhance the transparency, reproducibility, and completeness of the proposed benchmark and experimental findings.

\section{Crowdworking Details}
\label{sec:crowdworking}

\subsection{Processing and Filtering}\label{app:data-filter}

\begin{figure*}[tp]
    \centering
    \includegraphics[width=\linewidth]{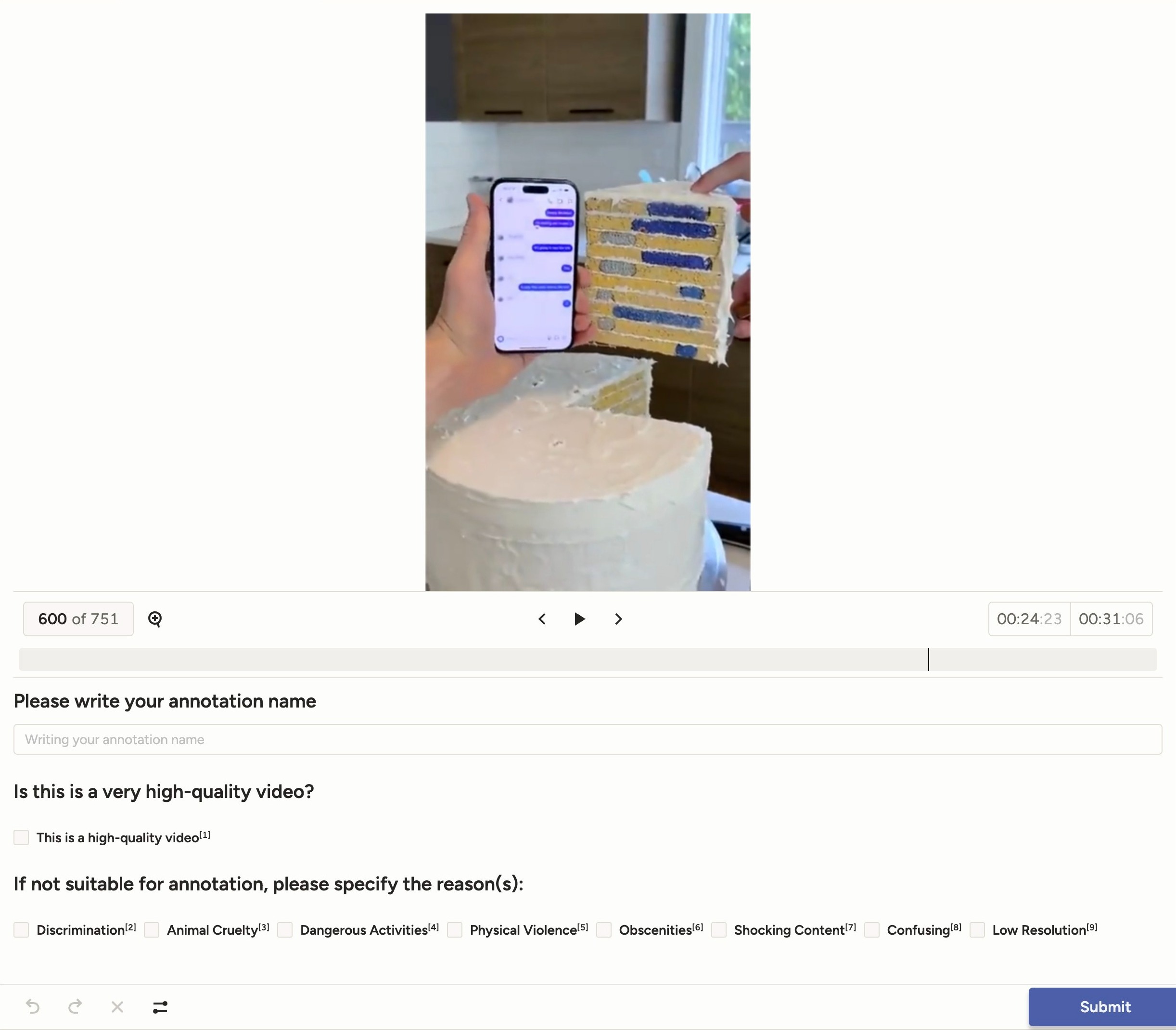}
    \caption{Interface for the Harmful Content Detection HIT.}
    \label{fig:CrownWorking_Fig1}
\end{figure*}

\paragraph{Harmful Content Detection.} Before the annotation process began, we manually filtered out videos that contained potentially harmful content to ensure the video data's safety and quality (Figure \ref{fig:CrownWorking_Fig1} visualizes our annotation interface). Based on the criteria outlined by \citet{thoppilan2022lamda}, we defined 6 categories of harmful contents, following aspects are checked for each video.
\begin{itemize}[leftmargin=*]
    \item \emph{Discrimination.} Videos that display discrimination based on race, gender, sexual orientation, age, disability, appearance (e.g., obesity), or religion.
    \item \emph{Animal Cruelty.} Videos that depict the abuse or mistreatment of animals.
    \item \emph{Dangerous Activities.} Videos that include dangerous content such as drug use, criminal behavior, bullying, terrorism, rumor propagation, incitement, or misinformation.
    \item \emph{Physical Violence.} Videos containing acts of physical violence against individuals, including fighting, severe injuries, bleeding, self-harm, or torture.
    \item \emph{Obscenities.} Videos that contain explicit language, sexual behavior, or suggestive content.
    \item \emph{Shocking Content.} Videos that include startling or fear-inducing elements such as gunshots, explosions, or jump scares.
\end{itemize}
In addition to harmful content detection, videos are also evaluated based on their quality:
\begin{itemize}[leftmargin=*]
    \item \emph{Confusing:} Videos that are incomplete or otherwise difficult to understand.
    \item \emph{Low Resolution:} Videos with a level of clarity that makes it challenging to discern the content.
\end{itemize}

\paragraph{Chaplin Video Segmentation.} 
We selected 62 silent films by Charlie Chaplin and hired annotators to meticulously review each film, manually recording humorous moments to ensure each mime clip illustrates a whole mime through a single event or multi events. And we removed videos where both the reason for the humor and the action were repetitive (e.g. humor arising from a comical action due to inflexibility, such as failing to position a ladder properly) to ensure the quality and consistency of the videos and their annotations.

\paragraph{Speech Reliance Minimization.} To ensure reliable identification of humorous content, we instructed two annotators to independently review each video and confirm the presence of clear humor. Each annotator was also instructed to review each video and label whether humor was primarily conveyed through visual cues and could be understood independently of speech. Only videos for which both annotators agreed were retained for the final dataset. We further employed Whisper~\citep{radford2023robust}, a performant speech-to-text model, to transcribe audio. Since Whisper transcribes filler sounds (e.g., “uh,” “hmm”) and other minimal utterances, we excluded any videos where the transcribed text exceeded 10 characters. Additionally, videos containing non-English speech were retained but muted, removing dependence on linguistic cues. 

\subsection{Annotation}\label{app:data-annotate}

\begin{figure*}[tp]
    \centering
    \includegraphics[width=\linewidth]{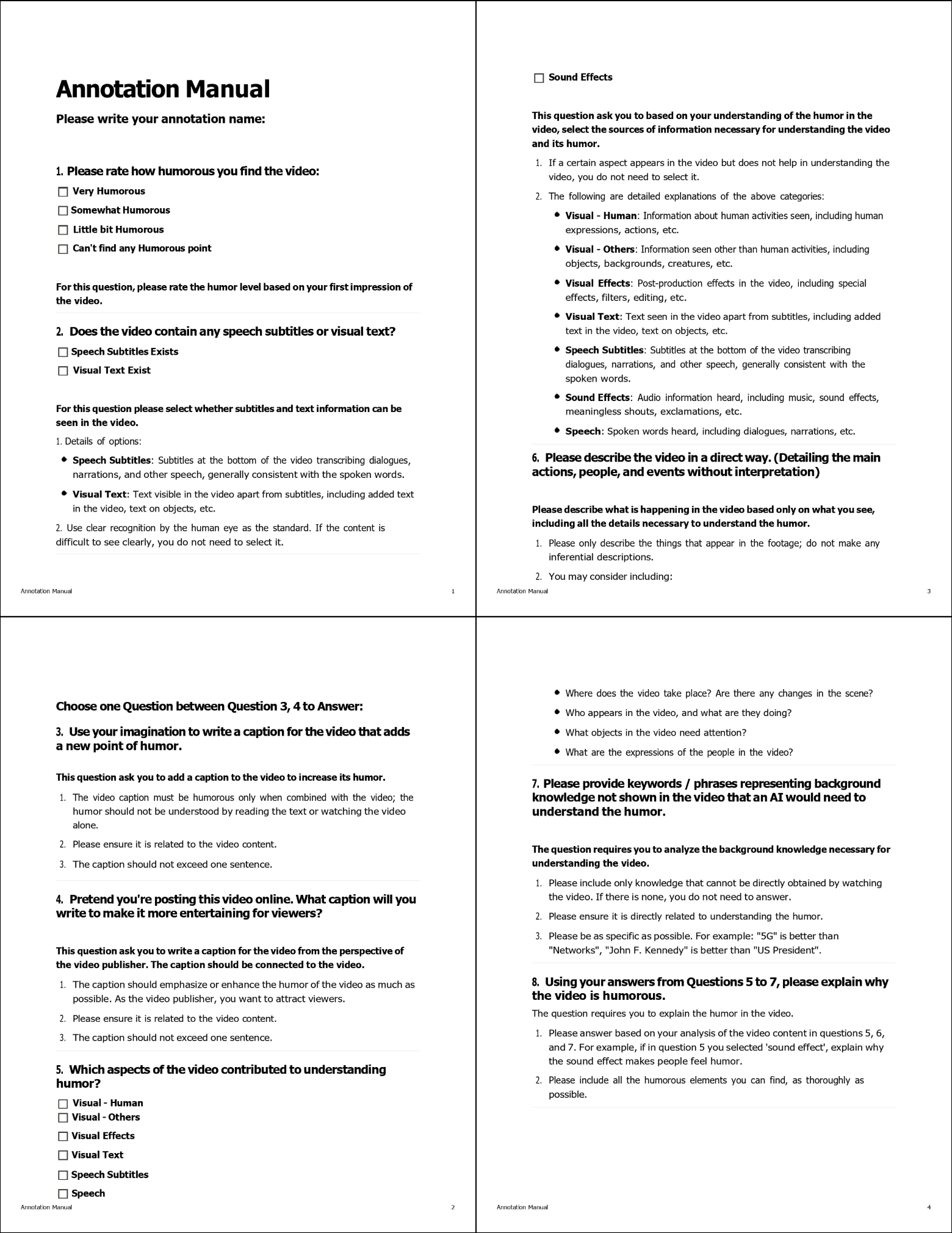}
    \caption{Interface for Annotation Manual for data annotation.}
    \label{fig:annotation_manual}
\end{figure*}

\textbf{Annotator Training.} We provided appropriate annotation training for crowdworkers, offering detailed explanations of the annotation platform’s usage and the annotation guidelines for different tasks. Additionally, we supplied an annotation manual (Figure \ref{fig:annotation_manual}) and corresponding instructional videos, which included specific descriptions and examples of the annotation requirements for crowdworkers to consult at any time during the annotation process.

\textbf{Qualification.} The recruited crowdworkers were mainly from China, all possessing at least an undergraduate education and with English background. Before formal annotation began, we conducted training sessions and a qualification review. During the qualification stage, crowdworkers were required to annotate 15 video samples. We manually reviewed their results and assigned scores based on the annotation guidelines. Ultimately, we selected eight qualified annotators. And we provided fair compensation to all crowdworkers, ensuring their hourly wages exceeded the local minimum wage.

\begin{figure}[ht!]
\begin{subfigure}[t]{\linewidth}
    \centering
    \includegraphics[width=\linewidth]{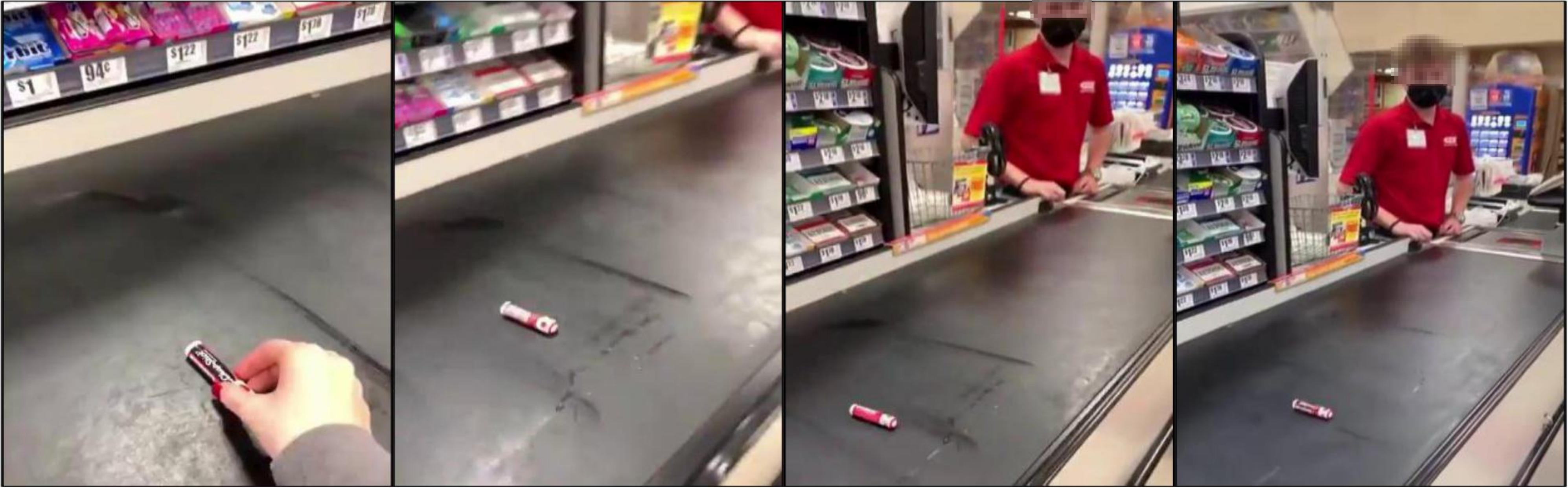}
    \caption[teaser-a]%
    {{\footnotesize \textbf{Visuals.} A man placed a battery on the conveyor belt, but it rolled against the belt's motion, forcing the cashier into an endless wait. For those who know \textcolor{mygreen}{the physics of a rolling cylinder on a moving conveyor}, the scene feels even more clever.
    }}    
    \label{fig:teaser-a}
\end{subfigure}
\hfill
\begin{subfigure}[t]{\linewidth}
    \centering
    \includegraphics[width=\linewidth]{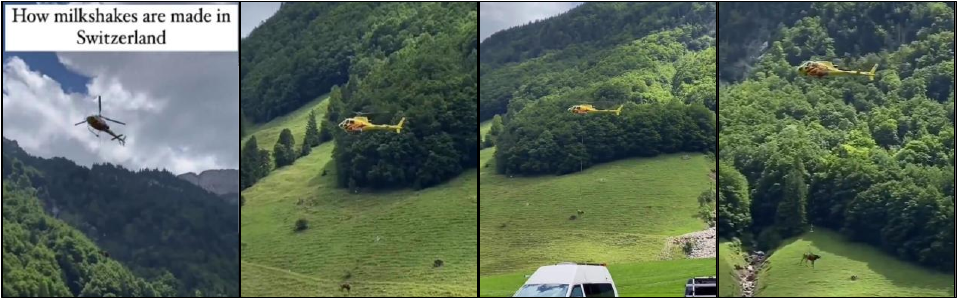}
    \caption[teaser-b]%
    {{\footnotesize \textbf{Visuals}+Text. The video shows an animal rescue, with \textcolor{mygreen}{a cow dangling beneath a helicopter}, appearing to swirl midair. The scene seems routine at first, but the added text  \textcolor{mygreen}{`{milkshakes}' cleverly parallels the moment}, making it unexpectedly witty.
    }}    
    \label{fig:teaser-b}
\end{subfigure}
\caption{\label{app_fig:teaser}
Examples of visual-centric humor understanding, where `text' refers to visual text.
}
\end{figure}

For the annotation process, we adopted a three-person collaborative annotation scheme, ensuring that each data entry underwent three rounds of annotation. First, an annotator performed the initial annotation. Next, a second annotator reviewed and supplemented the annotation. Finally, a third annotator reviewed and further refined the previous two rounds of annotations. The annotators rotated through these three roles, and each annotation round was tracked to ensure that the three rounds for each data entry were completed by different annotators. For humor rating and video captions, annotators were required to independently provide their own answers. For the remaining annotation tasks, when the second and third annotators reviewed and modified the previous annotations, they were required to submit a new annotation if they identified any issues. If a specific annotation issue was modified in all three rounds for a given video, we conducted a final review to assess the validity of the annotation results.

\subsection{Copyright \& License}
\label{app:copyright}
We respect the copyright of each video. We have emailed Charlie Chaplin's copyright holders regarding copyright issues related to Chaplin clips, and v-HUB is only used for academic research. Commercial use in any form is prohibited. The copyright of all videos belongs to the video owners, and we will remove the videos upon their request. Without prior approval, you cannot distribute, publish, copy, disseminate, or modify v-HUB in whole or in part. You must strictly comply with the above restrictions.


\begin{figure*}[htp]
    \centering
    \includegraphics[width=0.9\linewidth]{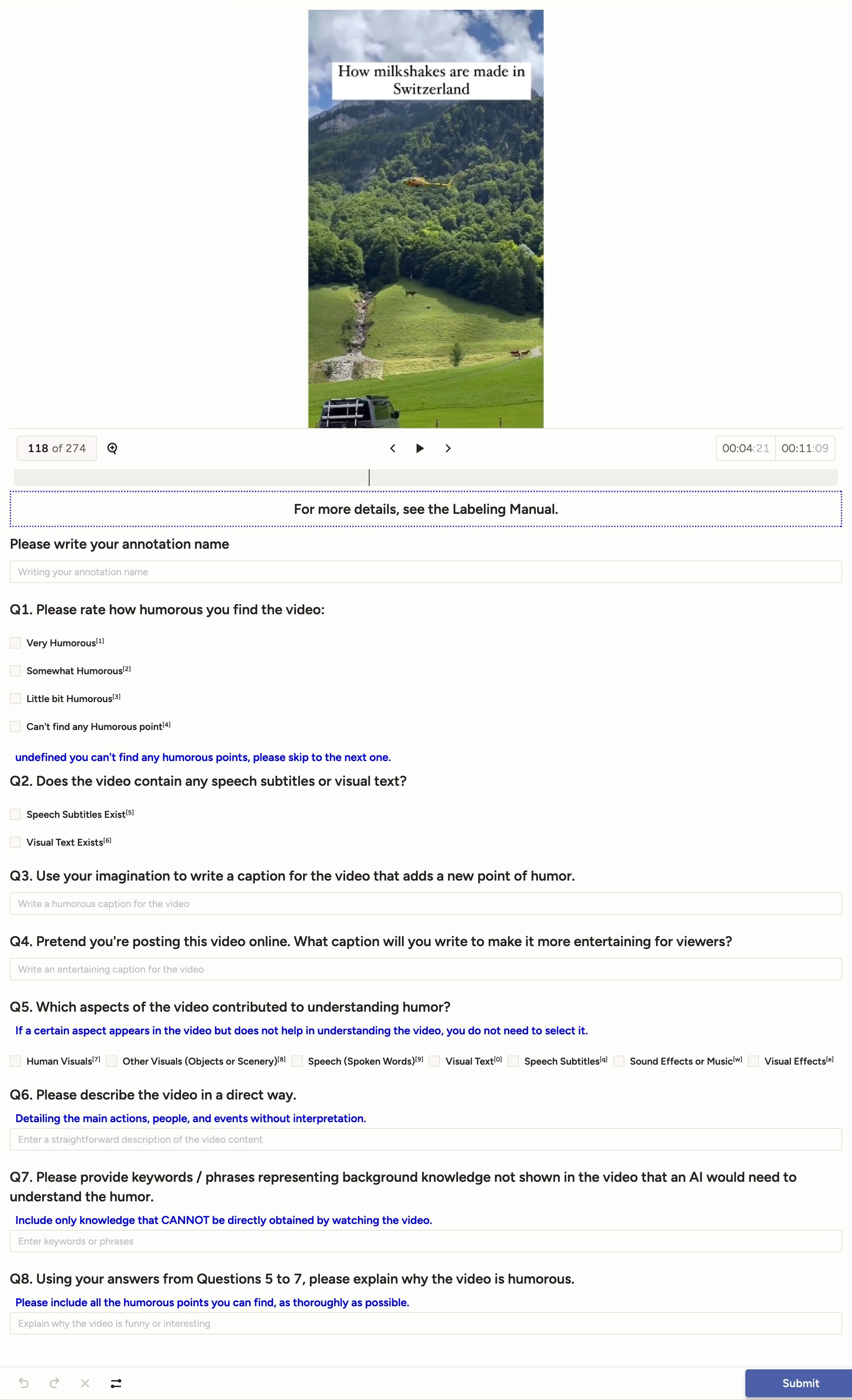}
    \caption{Interface for HIT.}
    \label{fig:crownWorking_fig2}
\end{figure*}

\section{Additional Experimental Notes}\label{app:add_exp_note}

\subsection{Details of Generate Open-ended QA Pairs}\label{app:generate_qapair}
We employed GPT-4o to generate QA pairs for each video, with the questions primarily covering temporal, descriptive, and causal aspects. The specific prompts used for QA generation are provided in Table~\ref{app_tab: prompt_for_open_ended_QA}. Subsequently, annotators manually reviewed and revised the QA pairs for each video to ensure their accuracy and quality.

\subsection{Details of Evaluation Methods}\label{app:eval-method}

We used the 3.9.1 NLTK, 0.3.13 bert-score, and 5.1.1 sentence-transformers package to calculate the metrics.

\paragraph{AutoDQ.} It evaluates the presence of humor-related events in the generated explanations~\citep{wang2024tarsier}. It extracts key events from the model’s output and compares them to ground truth (GT) annotations using entailment analysis. It provides three metrics: recall, precision, and F1 score, defined as:
\begin{itemize}[leftmargin=*]
    \item \emph{Recall} measures the percentage of GT events entailed by the model-generated events.
    \item \emph{Precision} measures the percentage of model-generated events that are entailed by GT events.
    \item \emph{F1 Score} is the harmonic mean of precision and recall, serving as a balanced indicator that jointly reflects coverage and correctness.
\end{itemize}
The inclusion of AutoDQ allows us to evaluate factual correctness and event coverage, checking whether the explanations cover all the humorous points in the video content.


\paragraph{Human Evaluation.} We randomly sampled 50 explanations generated by models for human evaluation. To ensure consistency in the evaluation criteria, we assigned one annotator to rate the humor explanations generated by models. The scores ranged from 0 to 100, and were subsequently normalized by a factor of 100, yielding final results within the range $[0,1]$.






\subsection{Baseline Models}
\begin{table}[]
\centering
\caption{Evaluated Models.}
\label{tab:model_list}
\resizebox{\linewidth}{!}{%
\begin{tabular}{@{}c|c|c|ccc@{}}
\toprule
\multirow{2}{*}{Models} & \multirow{2}{*}{\#Parameter} & \multirow{2}{*}{Proprietary} & \multicolumn{3}{c}{Input Modality}                                                \\ \cmidrule(l){4-6} 
                        &                            &                              & Text                      & Video                     & Video+Audio               \\ \midrule
Qwen2.5-VL              & 72B                        & \crossicon    & \checkicon & \checkicon & \crossicon \\
Qwen2.5-Omni    & 7B & \crossicon & \checkicon & \checkicon & \checkicon \\
Intern3.5-VL            & 8B                        & \crossicon    & \checkicon & \checkicon & \crossicon \\
MiniCPM2.6-o            & 8B                        & \crossicon    & \checkicon & \checkicon & \checkicon \\
Video-SALMONN-2  & 7B & \crossicon & \checkicon & \checkicon & \checkicon \\
GPT-4o                  & -                          & \checkicon    & \checkicon & \checkicon & \crossicon \\
Gemini-2.5-Flash & - & \checkicon & \checkicon & \checkicon & \checkicon \\ \bottomrule
\end{tabular}
}
\end{table}
To evaluate multimodal large language models’ ability to understand video humor, we selected state-of-the-art models representing three distinct input modalities, as summarized in Table~\ref{tab:model_list}. Specifically, we include  multimodal LLMs that process raw visual frames and text, and omni LLMs that integrate both text, video and audio signals. This set covers both public models (e.g., Qwen2.5-VL, Intern3.5-VL) and proprietary models (e.g., Gemini-2.5-Flash, GPT-4o), offering a broad perspective on current approaches. Each model is evaluated under all input conditions it can handle (see Section~\ref{sec:exp-setup}): for instance, omni-modal models can participate in the Text-Only, Video-Only, and Video+Audio groups, whereas multimodal models are tested exclusively with textual input and raw visual frames. This setup allows us to isolate how each model category—multimodal and omni-modal—contributes to humor understanding across diverse input modalities.

\section{Additional Experimental Results}\label{app:more-results}

\begin{table}[]
\centering
\caption{
The impact of requiring background knowledge support on video humor understanding in the Video-Only setting.
}
\label{tab:background_dependent}
\resizebox{\linewidth}{!}{%
\begin{tabular}{@{}c|ccc|c|cc@{}}
\toprule
\multicolumn{1}{c|}{\multirow{2}{*}{Models}} & \multicolumn{3}{c|}{Explanation}                & \multicolumn{1}{c|}{Matching} & \multicolumn{2}{c}{Open-ended QA} \\ \cmidrule(l){2-7} 
\multicolumn{1}{c|}{}                        & SentBERT & METEOR & \multicolumn{1}{l|}{AutoDQ} & \multicolumn{1}{c|}{Accuracy} & SentBERT         & METEOR         \\ \midrule
\multicolumn{7}{c}{\cellcolor[HTML]{EFEFEF}\textit{Background-Dependent videos}}                                                                                                                 \\ \midrule
\multicolumn{1}{c|}{Gemini-2.5-Flash}        & 0.500    & 0.211  & \multicolumn{1}{c|}{0.198}  & \multicolumn{1}{c|}{0.628}    & 0.433            & 0.266          \\
\multicolumn{1}{c|}{video-SALMONN-2}         & 0.271    & 0.153  & \multicolumn{1}{c|}{0.028}  & \multicolumn{1}{c|}{0.257}    & 0.307            & 0.160          \\
\multicolumn{1}{c|}{MiniCPM2.6-o}            & 0.402    & 0.166  & \multicolumn{1}{c|}{0.093}  & \multicolumn{1}{c|}{0.374}    & 0.328            & 0.120          \\
\multicolumn{1}{c|}{Qwen2.5-Omni}            & 0.401    & 0.158  & \multicolumn{1}{c|}{0.137}  & \multicolumn{1}{c|}{0.559}    & 0.382            & 0.105          \\ \midrule
\multicolumn{7}{c}{\cellcolor[HTML]{EFEFEF}\textit{Full dataset}}                                                                                                     \\ \midrule
\multicolumn{1}{c|}{Gemini-2.5-Flash}        & 0.459    & 0.199  & \multicolumn{1}{c|}{0.175}  & \multicolumn{1}{c|}{0.580}    & 0.424            & 0.270          \\
\multicolumn{1}{c|}{video-SALMONN-2}         & 0.269    & 0.150  & \multicolumn{1}{c|}{0.052}  & \multicolumn{1}{c|}{0.243}    & 0.317            & 0.169          \\
\multicolumn{1}{c|}{MiniCPM2.6-o}            & 0.381    & 0.165  & \multicolumn{1}{c|}{0.112}  & \multicolumn{1}{c|}{0.362}    & 0.369            & 0.186          \\
\multicolumn{1}{c|}{Qwen2.5-Omni}            & 0.384    & 0.159  & \multicolumn{1}{c|}{0.144}  & \multicolumn{1}{c|}{0.553}    & 0.382            & 0.121          \\ \bottomrule
\end{tabular}
}
\vspace{-0.8em}
\end{table}


\paragraph{Comparable performance on videos with and without background knowledge requirements.} The results in Table~\ref{tab:background_dependent} show that model performance on Background-Dependent videos is largely comparable to that on the full dataset in the Video-Only setting. For example, video-SALMONN-2 attains an average SentBERT of 0.271 on the humor explanation task for background-dependent videos, which is statistically similar to its SentBERT of 0.269 on the full dataset. This suggests that the language-model component of MLLMs already encodes most of the cultural background knowledge necessary for humor comprehension, meaning that \emph{the absence of explicit background knowledge in the input does not significantly degrade their performance}. MLLMs show a comparable performance in understanding videos that require background knowledge compared to those that do not, potentially because video humor rarely relies on specific background knowledge, making it universally understandable.


\paragraph{Proprietary MLLMs show stronger resilience to multimodal inputs compared to public MLLMs.} 

The results in Table~\ref{tab:human_preference} indicate that current MLLMs rely heavily on linguistic cues to generate reasonable explanations, and struggle to effectively extract semantic information from raw visual or auditory signals. For example, Qwen2.5-VL attains a preference score of 0.687 under Text-Only, significantly outperforming its Video-Only score of 0.423. Furthermore, although closed-source models demonstrate greater robustness under multimodal inputs, they still struggle to align visual and audio cues to enhance humor comprehension. For instance, Gemini-2.5-Flash achieves 0.546 (Video-Only) and 0.566 (Video+Audio).

\section{Case Study}
\label{sec:case_study}
\begin{figure*}[tp]
    \centering
    \begin{subfigure}[b]{0.78\linewidth}
        \centering
        \includegraphics[width=\linewidth]{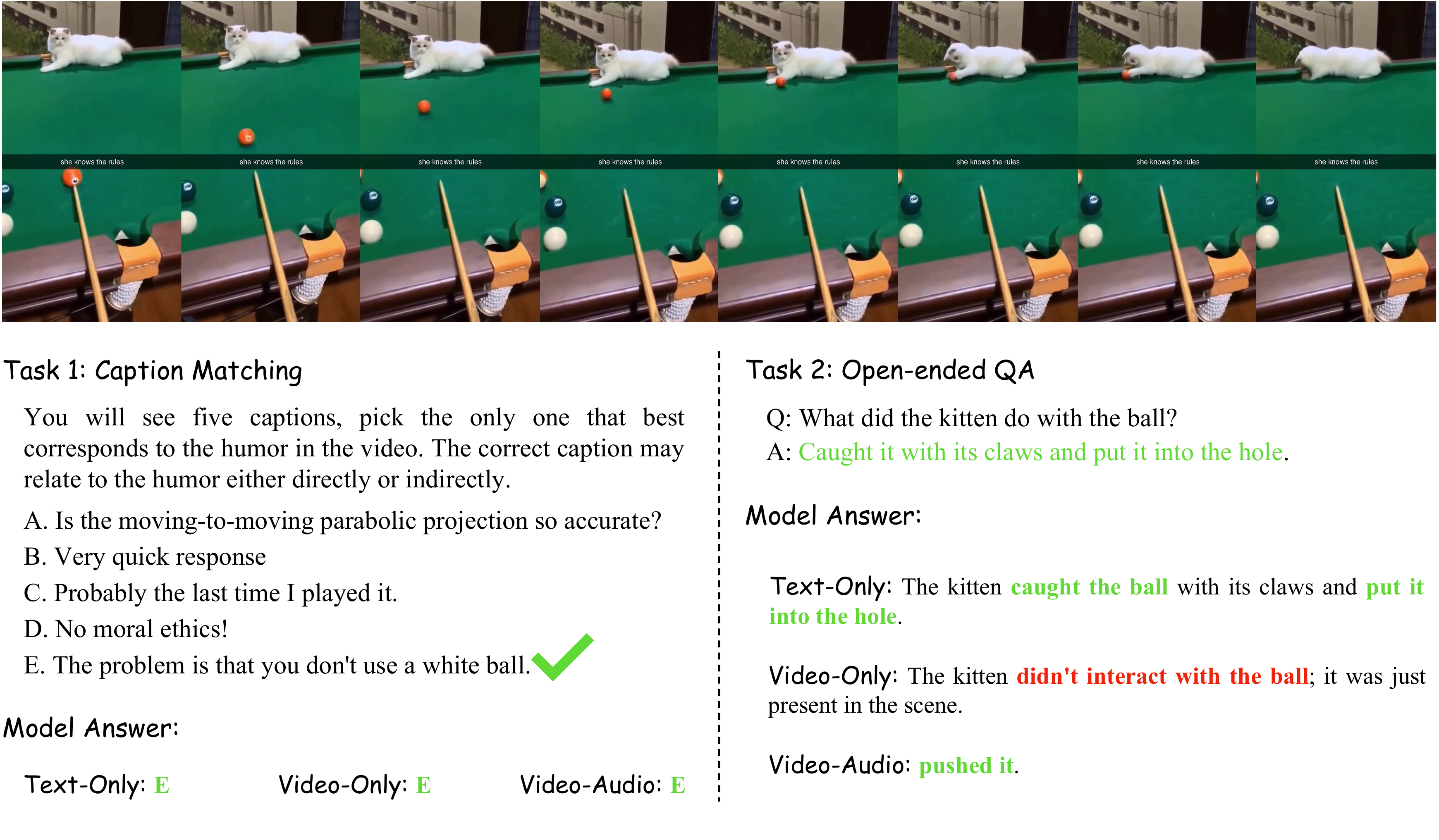}
        \caption{case study 1}
        \label{fig:case1}
    \end{subfigure}
    
    \vspace{1em} 
    
    \begin{subfigure}[b]{0.78\linewidth}
        \centering
        \includegraphics[width=\linewidth]{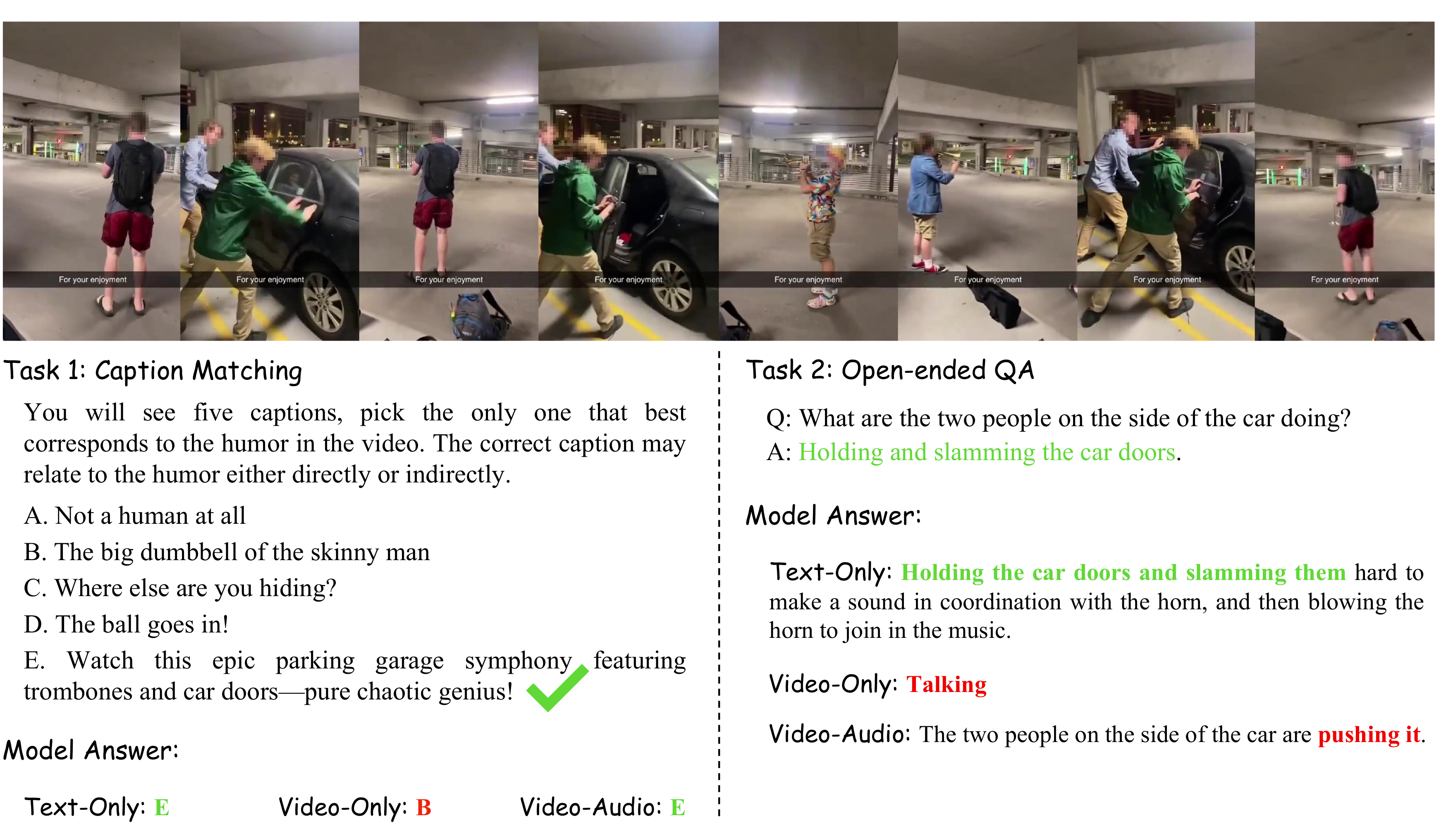}
        \caption{case study 2}
        \label{fig:case2}
    \end{subfigure}
    
    \vspace{1em}
    
    \begin{subfigure}[b]{0.78\linewidth}
        \centering
        \includegraphics[width=\linewidth]{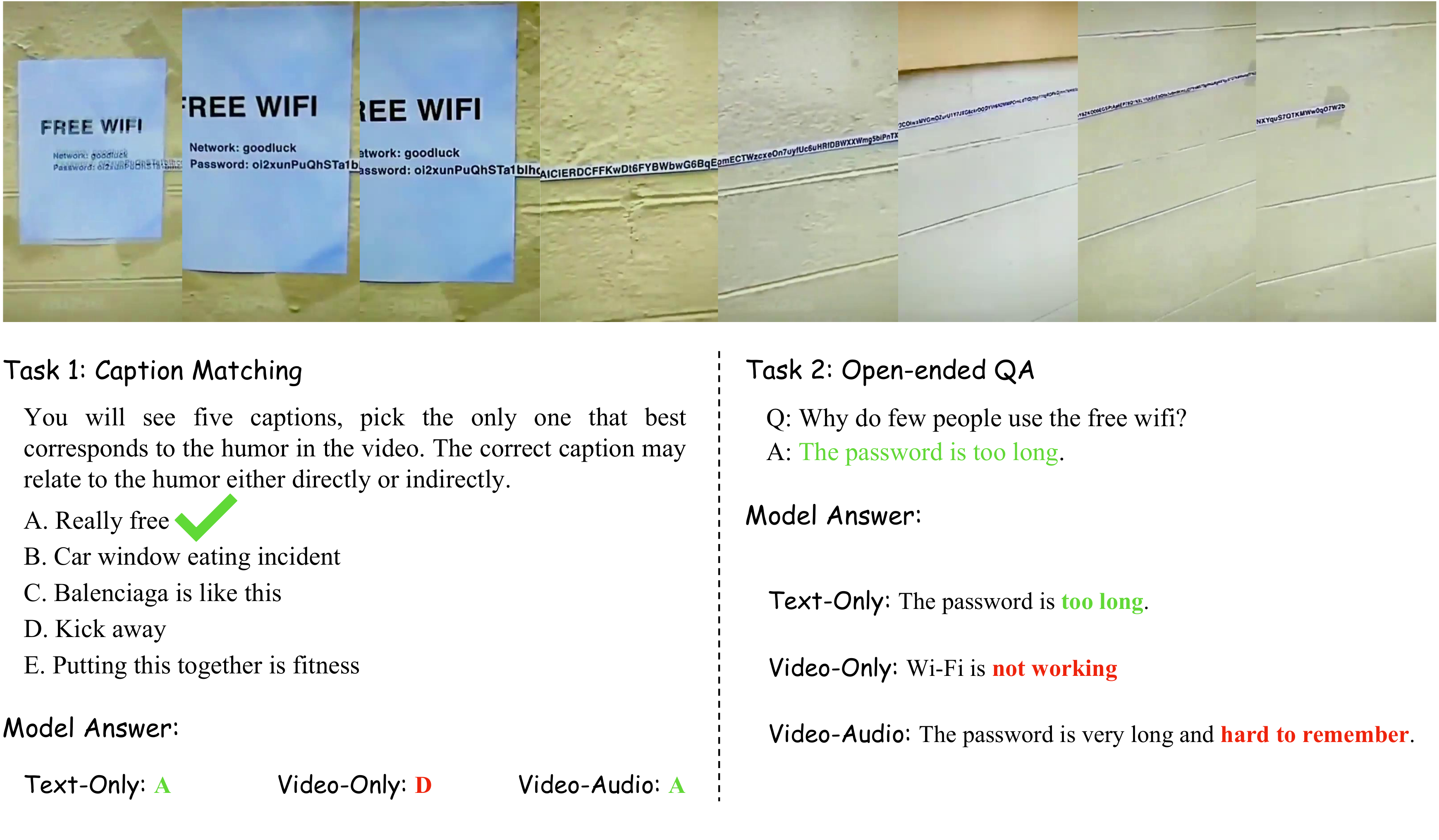}
        \caption{case study 3}
        \label{fig:case3}
    \end{subfigure}
    \caption{Three case studies illustrating our findings.}
    \label{fig:case study}
\end{figure*}

We present our case studies in Figure~\ref{fig:case study}.
\begin{table}[]
\centering
\caption{
Human preference comparison of humor explanations across four model categories.}
\label{tab:human_preference}
\resizebox{\linewidth}{!}{%
\begin{tabular}{@{}c|c|c|ccc@{}}
\toprule
\multirow{2}{*}{Models} & \multirow{2}{*}{Proprietary} & \multirow{2}{*}{Type} & \multicolumn{3}{c}{Setting} \\ \cmidrule(l){4-6} 
                 &                           &         & Text-Only & Video-Only & Video+Audio \\ \midrule
Qwen2.5-VL   & \crossicon & MLLM    & 0.687            & 0.423      & --               \\
Qwen2.5-Omni     & \crossicon & OmniLLM & 0.574            & 0.430      & 0.381            \\
GPT-4o           & \checkicon & MLLM    & 0.654            & 0.576      & --               \\
Gemini-2.5-Flash & \checkicon & OmniLLM & 0.651            & 0.546      & 0.566            \\ \bottomrule
\end{tabular}
}
\vspace{-1.5em}
\end{table}

\begin{table*}[ht]
\centering
\caption{
The impact of visual text on video humor understanding in the Video+Audio setting.}
\label{add_tab:visual_sound}
\resizebox{\linewidth}{!}{%
\begin{tabular}{@{}ccccllcccccllcc@{}}
\toprule
\multicolumn{1}{c|}{}                         & \multicolumn{7}{c|}{Sound contributing to humor}                                                                                                                & \multicolumn{7}{c}{Sound not contributing to humor}                                                                                        \\ \cmidrule(l){2-15} 
\multicolumn{1}{c|}{}                         & \multicolumn{6}{c|}{Explanation}                                                                                           & \multicolumn{1}{c|}{Open-ended QA} & \multicolumn{6}{c|}{Explanation}                                                                                           & Open-ended QA \\ \cmidrule(l){2-15} 
\multicolumn{1}{c|}{\multirow{-3}{*}{Models}} & \multicolumn{1}{l}{SentBERT} & \multicolumn{1}{l}{METEOR} & BERTScore & Precision & Recall & \multicolumn{1}{c|}{F1 Score} & \multicolumn{1}{c|}{BERTScore}     & \multicolumn{1}{l}{SentBERT} & \multicolumn{1}{l}{METEOR} & BERTScore & Precision & Recall & \multicolumn{1}{c|}{F1 Score} & BERTScore     \\ \midrule
\multicolumn{15}{c}{\cellcolor[HTML]{EFEFEF}\textit{w/ visual text}}                                                                                                                                                                                                                                                                                                  \\ \midrule
\multicolumn{1}{c|}{Gemini-2.5-Flash}         & 0.458                        & 0.199                      & 0.541     & 0.157     & 0.210  & \multicolumn{1}{c|}{0.174}    & \multicolumn{1}{c|}{0.552}         & 0.472                        & 0.206                      & 0.547     & 0.185     & 0.174  & \multicolumn{1}{c|}{0.177}    & 0.550         \\
\multicolumn{1}{c|}{video-SALMONN-2}          & 0.258                        & 0.164                      & 0.481     & 0.169     & 0.042  & \multicolumn{1}{c|}{0.067}    & \multicolumn{1}{c|}{0.532}         & 0.261                        & 0.160                      & 0.509     & 0.077     & 0.045  & \multicolumn{1}{c|}{0.057}    & 0.540         \\
\multicolumn{1}{c|}{MiniCPM2.6-o}             & 0.416                        & 0.170                      & 0.516     & 0.105     & 0.111  & \multicolumn{1}{c|}{0.108}    & \multicolumn{1}{c|}{0.527}         & 0.434                        & 0.182                      & 0.522     & 0.143     & 0.137  & \multicolumn{1}{c|}{0.151}    & 0.536         \\
\multicolumn{1}{c|}{Qwen2.5-Omni}            & 0.441                        & 0.170                      & 0.522     & 0.148     & 0.133  & \multicolumn{1}{c|}{0.140}    & \multicolumn{1}{c|}{0.521}         & 0.430                        & 0.176                      & 0.522     & 0.159     & 0.133  & \multicolumn{1}{c|}{0.145}    & 0.538         \\ \midrule
\multicolumn{15}{c}{\cellcolor[HTML]{EFEFEF}\textit{w/o visual text}}                                                                                                                                                                                                                                                                                                 \\ \midrule
\multicolumn{1}{c|}{Gemini-2.5-Flash}         & 0.482                        & 0.202                      & 0.550     & 0.165     & 0.195  & \multicolumn{1}{c|}{0.179}    & \multicolumn{1}{c|}{0.539}         & 0.448                        & 0.195                      & 0.543     & 0.034     & 0.051  & \multicolumn{1}{c|}{0.041}    & 0.552         \\
\multicolumn{1}{c|}{video-SALMONN-2}          & 0.276                        & 0.185                      & 0.496     & 0.124     & 0.045  & \multicolumn{1}{c|}{0.064}    & \multicolumn{1}{c|}{0.529}         & 0.285                        & 0.170                      & 0.504     & 0.098     & 0.034  & \multicolumn{1}{c|}{0.050}    & 0.541         \\
\multicolumn{1}{c|}{MiniCPM2.6-o}             & 0.407                        & 0.182                      & 0.525     & 0.130     & 0.097  & \multicolumn{1}{c|}{0.103}    & \multicolumn{1}{c|}{0.515}         & 0.364                        & 0.167                      & 0.510     & 0.060     & 0.048  & \multicolumn{1}{c|}{0.046}    & 0.514         \\
\multicolumn{1}{c|}{Qwen2.5-Omni}            & 0.425                        & 0.181                      & 0.538     & 0.116     & 0.128  & \multicolumn{1}{c|}{0.122}    & \multicolumn{1}{c|}{0.509}         & 0.413                        & 0.173                      & 0.522     & 0.062     & 0.045  & \multicolumn{1}{c|}{0.050}    & 0.529         \\ \bottomrule
\end{tabular}
}
\end{table*}

\begin{table*}[]
\centering
\caption{Comparison between MLLMs and their base LLMs under the Text-Only setting.}
\label{tab:add_mllmvsllm}
\resizebox{\linewidth}{!}{%
\begin{tabular}{@{}c|cccccc|c@{}}
\toprule
\multirow{2}{*}{Models}              & \multicolumn{6}{c|}{Explanation}                                                                        & Matching \\ \cmidrule(l){2-8} 
                                     & SentBERT & METEOR & \multicolumn{1}{l}{BERTScore} & Precision & Recall & \multicolumn{1}{c|}{F1 Score} & Accuracy \\ \midrule
\multicolumn{1}{c|}{Qwen2.5-VL}  & 0.543    & 0.250  & 0.573                         & 0.323     & 0.364  & \multicolumn{1}{c|}{0.342}    & 0.719    \\
\multicolumn{1}{c|}{Qwen2.5-72B}     & 0.536    & 0.245  & 0.576                         & 0.318     & 0.374  & \multicolumn{1}{c|}{0.344}    & 0.661    \\
\multicolumn{1}{c|}{Qwen2.5-Omni} & 0.536    & 0.233  & 0.565                         & 0.303     & 0.331  & \multicolumn{1}{c|}{0.316}    & 0.644    \\
\multicolumn{1}{c|}{Qwen2.5-7B}      & 0.560    & 0.240  & 0.567                         & 0.293     & 0.369  & \multicolumn{1}{c|}{0.324}    & 0.542    \\ \bottomrule
\end{tabular}
}
\end{table*}

\begin{table}[ht]
\centering
\caption{Model performance on Humor Explanation.}
\label{tab:add-main-result}
\resizebox{\linewidth}{!}{%
\begin{tabular}{@{}cccccc@{}}
\toprule
\multicolumn{1}{c|}{}                         & \multicolumn{4}{c|}{Explanation}                                                   & \multicolumn{1}{l}{Open-ended QA} \\ \cmidrule(l){2-6} 
\multicolumn{1}{c|}{\multirow{-2}{*}{Models}} & \multicolumn{1}{l}{BERTScore} & Precision & Recall & \multicolumn{1}{c|}{F1 Score} & BERTScore                         \\ \midrule
\multicolumn{6}{c}{\cellcolor[HTML]{EFEFEF}\textit{Text-Only}}                                                                                                                  \\ \midrule
\multicolumn{1}{c|}{Gemini-2.5-Flash}         & 0.575                         & 0.307     & 0.385  & \multicolumn{1}{c|}{0.342}    & 0.712                             \\
\multicolumn{1}{c|}{video-SALMONN-2}          & 0.586                         & 0.290     & 0.350  & \multicolumn{1}{c|}{0.317}    & 0.639                             \\
\multicolumn{1}{c|}{MiniCPM2.6-o}             & 0.558                         & 0.307     & 0.345  & \multicolumn{1}{c|}{0.325}    & 0.536                             \\
\multicolumn{1}{c|}{Qwen2.5-Omni}            & 0.565                         & 0.303     & 0.331  & \multicolumn{1}{c|}{0.316}    & 0.687                             \\
\multicolumn{1}{c|}{Qwen2.5-VL}          & 0.573                         & 0.323     & 0.364  & \multicolumn{1}{c|}{0.342}    & 0.730                             \\
\multicolumn{1}{c|}{Intern3.5-VL}             & 0.576                         & 0.320     & 0.382  & \multicolumn{1}{c|}{0.348}    & 0.685                             \\
\multicolumn{1}{c|}{GPT-4o}                   & 0.574                         & 0.339     & 0.417  & \multicolumn{1}{c|}{0.374}    & 0.699                             \\ \midrule
\multicolumn{6}{c}{\cellcolor[HTML]{EFEFEF}\textit{Video-Only}}                                                                                                                 \\ \midrule
\multicolumn{1}{c|}{Gemini-2.5-Flash}         & 0.546                         & 0.154     & 0.206  & \multicolumn{1}{c|}{0.176}    & 0.549                             \\
\multicolumn{1}{c|}{video-SALMONN-2}          & 0.497                         & 0.087     & 0.042  & \multicolumn{1}{c|}{0.052}    & 0.525                             \\
\multicolumn{1}{c|}{MiniCPM2.6-o}             & 0.516                         & 0.116     & 0.109  & \multicolumn{1}{c|}{0.112}    & 0.469                             \\
\multicolumn{1}{c|}{Qwen2.5-Omni}            & 0.518                         & 0.169     & 0.126  & \multicolumn{1}{c|}{0.144}    & 0.497                             \\
\multicolumn{1}{c|}{Qwen2.5-VL}          & 0.542                         & 0.150     & 0.152  & \multicolumn{1}{c|}{0.150}    & 0.546                             \\
\multicolumn{1}{c|}{Intern3.5-VL}             & 0.537                         & 0.126     & 0.116  & \multicolumn{1}{c|}{0.125}    & 0.540                             \\
\multicolumn{1}{c|}{GPT-4o}                   & 0.536                         & 0.214     & 0.205  & \multicolumn{1}{c|}{0.206}    & 0.544                             \\ \midrule
\multicolumn{6}{c}{\cellcolor[HTML]{EFEFEF}\textit{Video+Sound}}                                                                                                                \\ \midrule
\multicolumn{1}{c|}{Gemini-2.5-Flash}         & 0.546                         & 0.153     & 0.200  & \multicolumn{1}{c|}{0.173}    & 0.549                             \\
\multicolumn{1}{c|}{video-SALMONN-2}          & 0.499                         & 0.126     & 0.047  & \multicolumn{1}{c|}{0.066}    & 0.537                             \\
\multicolumn{1}{c|}{MiniCPM2.6-o}             & 0.519                         & 0.122     & 0.062  & \multicolumn{1}{c|}{0.110}    & 0.514                             \\
\multicolumn{1}{c|}{Qwen2.5-Omni}            & 0.525                         & 0.137     & 0.116  & \multicolumn{1}{c|}{0.120}    & 0.529                             \\ \bottomrule
\end{tabular}
}
\end{table}

\begin{table}[]
\centering
\caption{The impact of requiring background knowledge support on video humor understanding in the Video-Only setting.}
\resizebox{\linewidth}{!}{%
\begin{tabular}{@{}c|cccc|c@{}}
\toprule
\multicolumn{1}{c|}{\multirow{2}{*}{Models}} & \multicolumn{4}{c|}{Explanation}                                                   & \multicolumn{1}{l}{Open-ended QA} \\ \cmidrule(l){2-6} 
\multicolumn{1}{c|}{}                        & \multicolumn{1}{l}{BERTScore} & Precision & Recall & \multicolumn{1}{c|}{F1 Score} & BERTScore                         \\ \midrule
\multicolumn{6}{c}{\cellcolor[HTML]{EFEFEF}\textit{Background-Dependent videos}}                                                                                                                    \\ \midrule
\multicolumn{1}{c|}{Gemini-2.5-Flash}        & 0.566                         & 0.193     & 0.203  & \multicolumn{1}{c|}{0.198}    & 0.549                             \\
\multicolumn{1}{c|}{video-SALMONN-2}         & 0.512                         & 0.051     & 0.019  & \multicolumn{1}{c|}{0.028}    & 0.488                             \\
\multicolumn{1}{c|}{MiniCPM2.6-o}            & 0.531                         & 0.109     & 0.079  & \multicolumn{1}{c|}{0.093}    & 0.446                             \\
\multicolumn{1}{c|}{Qwen2.5-Omni}            & 0.530                         & 0.184     & 0.109  & \multicolumn{1}{c|}{0.137}    & 0.519                             \\ \midrule
\multicolumn{6}{c}{\cellcolor[HTML]{EFEFEF}\textit{Full dataset}}                                                                                                        \\ \midrule
\multicolumn{1}{c|}{Gemini-2.5-Flash}        & 0.546                         & 0.154     & 0.206  & \multicolumn{1}{c|}{0.176}    & 0.549                             \\
\multicolumn{1}{c|}{video-SALMONN-2}         & 0.497                         & 0.087     & 0.042  & \multicolumn{1}{c|}{0.052}    & 0.525                             \\
\multicolumn{1}{c|}{MiniCPM2.6-o}            & 0.516                         & 0.116     & 0.109  & \multicolumn{1}{c|}{0.112}    & 0.469                             \\
\multicolumn{1}{c|}{Qwen2.5-Omni}            & 0.518                         & 0.169     & 0.126  & \multicolumn{1}{c|}{0.144}    & 0.497                             \\ \bottomrule
\end{tabular}
}
\end{table}

\begin{table}[ht]
\centering
\caption{The impact of background knowledge on video humor understanding in the Video+Audio setting.}
\label{tab:add_with_background}
\resizebox{\linewidth}{!}{%
\begin{tabular}{@{}cccccc@{}}
\toprule
\multicolumn{1}{c|}{}                         & \multicolumn{4}{c|}{Explanation}                                                   & \multicolumn{1}{l}{Open-ended QA} \\ \cmidrule(l){2-6} 
\multicolumn{1}{c|}{\multirow{-2}{*}{Models}} & \multicolumn{1}{l}{BERTScore} & Precision & Recall & \multicolumn{1}{c|}{F1 Score} & BERTScore                         \\ \midrule
\multicolumn{6}{c}{\cellcolor[HTML]{EFEFEF}\textit{w/ Background Knowledge}}                                                                                                    \\ \midrule
\multicolumn{1}{c|}{video-SALMONN-2}          & 0.562                         & 0.117     & 0.107  & \multicolumn{1}{c|}{0.114}    & 0.536                             \\
\multicolumn{1}{c|}{MiniCPM2.6-o}            & 0.555                         & 0.192     & 0.194  & \multicolumn{1}{c|}{0.193}    & 0.520                             \\
\multicolumn{1}{c|}{Qwen2.5-Omni}            & 0.557                         & 0.195     & 0.160  & \multicolumn{1}{c|}{0.176}    & 0.555                             \\ \midrule
\multicolumn{6}{c}{\cellcolor[HTML]{EFEFEF}\textit{w/o Background Knowledge}}                                                                                                   \\ \midrule
\multicolumn{1}{c|}{video-SALMONN-2}          & 0.514                         & 0.084     & 0.014  & \multicolumn{1}{c|}{0.025}    & 0.528                             \\
\multicolumn{1}{c|}{MiniCPM2.6-o}            & 0.538                         & 0.132     & 0.103  & \multicolumn{1}{c|}{0.115}    & 0.509                             \\
\multicolumn{1}{c|}{Qwen2.5-Omni}            & 0.544                         & 0.142     & 0.114  & \multicolumn{1}{c|}{0.127}    & 0.525                             \\ \bottomrule
\end{tabular}
}
\end{table}

\begin{table}[ht]
\centering
\caption{
The impact of video era on video humor understanding in the Video-Only setting.
}
\label{tab:add_ccsf-vs-ugfv}
\resizebox{\linewidth}{!}{%
\begin{tabular}{@{}cccccc@{}}
\toprule
\multicolumn{1}{c|}{\multirow{2}{*}{Models}} & \multicolumn{4}{c|}{Explanation}                                                   & \multicolumn{1}{l}{Open-ended QA} \\ \cmidrule(l){2-6}  
\multicolumn{1}{c|}{}                        & \multicolumn{1}{l}{BERTScore} & Precision & Recall & \multicolumn{1}{c|}{F1 Score} & BERTScore                         \\ \midrule
\multicolumn{6}{c}{\cellcolor[HTML]{EFEFEF}\textit{\textbf{Last-Century} Charlie Chaplin's Silent Films}}                                                                                                       \\ \midrule
\multicolumn{1}{c|}{Gemini-2.5-Flash}        & 0.541                         & 0.118     & 0.145  & \multicolumn{1}{c|}{0.130}    & 0.545                             \\
\multicolumn{1}{c|}{video-SALMONN-2}         & 0.509                         & 0.035     & 0.007  & \multicolumn{1}{c|}{0.012}    & 0.513                             \\
\multicolumn{1}{c|}{MiniCPM2.6-o}            & 0.508                         & 0.116     & 0.083  & \multicolumn{1}{c|}{0.097}    & 0.470                             \\
\multicolumn{1}{c|}{Qwen2.5-Omni}            & 0.510                         & 0.153     & 0.070  & \multicolumn{1}{c|}{0.096}    & 0.493                             \\ \midrule
\multicolumn{6}{c}{\cellcolor[HTML]{EFEFEF}\textit{\textbf{Contemporary} User-Generated Funny Video}}                                                                                                           \\ \hline
\multicolumn{1}{c|}{Gemini-2.5-Flash}        & 0.547                         & 0.173     & 0.222  & \multicolumn{1}{c|}{0.194}    & 0.550                             \\
\multicolumn{1}{c|}{video-SALMONN-2}         & 0.511                         & 0.103     & 0.052  & \multicolumn{1}{c|}{0.061}    & 0.530                             \\
\multicolumn{1}{c|}{MiniCPM2.6-o}            & 0.518                         & 0.113     & 0.124  & \multicolumn{1}{c|}{0.118}    & 0.470                             \\
\multicolumn{1}{c|}{Qwen2.5-Omni}            & 0.520                         & 0.173     & 0.130  & \multicolumn{1}{c|}{0.166}    & 0.498                             \\ \bottomrule
\end{tabular}
}
\end{table}

\section{Prompts}
\label{sec:prompts}
We list our prompt in \Cref{app_tab: prompt_writing_caption_of_videos,app_tab: prompt_for_open_ended_QA,app_tab: prompt_for_video_caption_matching,app_tab: prompt_for_video_dec_qa,app_tab: prompt_for_video_description_caption_matching,app_tab: prompt_for_video_description_explanation,app_tab: prompt_for_video_exp,app_tab: prompt_for_video_qa,app_tab: prompt_for_video_with_sound_caption_matching,app_tab: prompt_for_video_with_sound_exp,app_tab: prompt_for_video_with_sound_QA,app_tab: prompt_for_video_with_background_knowledge_caption_matching}

\begin{table*}[h!]
\centering
\caption{Prompt for generate QA pairs.}

\begin{tabular}{p{0.96\linewidth}}
\toprule
\tt These are frames from a video. \\
\tt And you'll be given a description of a video and an explanation of why it's humorous to watch. \\
\tt Based on given information, generate a Video Reasoning QA pair, try to make answer only as phrases. Let’s think step by step.\ \textbackslash n \\
\tt Additionally, classify this question into one of the following categories using the concise definitions provided:\ \textbackslash n \\
\tt Descriptive question: Involves factual details such as location or count\ \textbackslash n \\
\tt Temporal question: Involves time-related aspects (e.g., previous, after)\ \textbackslash n \\
\tt Causal question: Involves reasons or explanations (e.g., why, how)\ \textbackslash n\textbackslash n \\
\tt Example 1:\ \textbackslash n \\
\tt Description:\ \textbackslash n \\
\tt Two hands are stretched out, one hand holding KFC chicken nuggets and the other hand holding seeds. In the distance, a chicken runs over, but the chicken prefers to eat the KFC chicken.\ \textbackslash n \\
\tt Explanation: The chicken surprisingly likes to eat KFC chicken, which is unexpected and a bit funny. The man realizes something is wrong and tries to push the chicken pieces away with his hand, which adds to the humor with a sense of panic.\ \textbackslash n\textbackslash n \\
\tt Question: What does the man holding in his hand?\ \textbackslash n \\
\tt Answer: KFC chicken nuggets and seeds.\ \textbackslash n \\
\tt Type: Descriptive\ \textbackslash n\textbackslash n \\
\tt Example 2:\ \textbackslash n \\
\tt Description: A man poured red liquid into the water, and a group of fish came to snatch the food. Another man poured beer into the water, and a group of men came to snatch the food like fish.\ \textbackslash n \\
\tt Explanation: The portrait of people snatching food like fish humorously reflects the attraction of beer to men, and the connection between them is very funny.\ \textbackslash n\textbackslash n \\
\tt Question: After the man poured beer into the water, what happened?\ \textbackslash n \\
\tt Answer: A group of men came.\ \textbackslash n \\
\tt Type: Temporal\ \textbackslash n\textbackslash n \\
\tt Example 3:\ \textbackslash n \\
\tt Description: A woman was lying on the handrail of an escalator while moving down. A man saw her, and lying on the handrail on the other side, and as a result, there was no barrier on that side, and he fell directly down the escalator.\ \textbackslash n \\
\tt Explanation: The man tried to show off by imitating others, but ended up falling hard, which made people find it funny.\ \textbackslash n\textbackslash n \\
\tt Question: Why does the man fall off on the other side of the handrail?\ \textbackslash n \\
\tt Answer: There was no barrier.\ \textbackslash n \\
\tt Type: Causal\ \textbackslash n\textbackslash n \\
\tt Output format:\ \textbackslash n \\
\tt Question: \textless question\textgreater\ \textbackslash n \\
\tt Answer: \textless answer\textgreater\ \textbackslash n \\
\tt Type: \textless type\textgreater\ \textbackslash n\textbackslash n \\
\tt Video Description: \{video\_description\}\ \textbackslash n \\
\tt Humor Explanation: \{humor\_explanation\}\ \textbackslash n \\
\bottomrule
\end{tabular}
\label{app_tab: prompt_for_open_ended_QA}

\end{table*}

\begin{table}[h!]
\centering
\caption{Prompt for video QA.}

\begin{tabular}{p{0.96\linewidth}}
\toprule
\tt System: You are a helpful AI assistant. You can analyze videos and answer questions about their content. Respond with short and concise answers. Avoid using unpronouncable punctuation or emojis. \\
\tt \\
\tt User: These are frames from a video. Based on these frames, answer the following question: \{question\}\ \textbackslash n\textbackslash n \\
\tt \\
\tt Output format:\ \textbackslash n \\
\tt Answer: \textless answer\textgreater\ \textbackslash n\textbackslash n \\
\bottomrule
\end{tabular}

\label{app_tab: prompt_for_video_qa}

\end{table}

\begin{table}[h!]
\centering
\caption{Prompt for video explanation.}

\begin{tabular}{p{0.96\linewidth}}
\toprule
\tt System: You are a helpful AI assistant specialized in video understanding and humor analysis. You can explain jokes clearly and naturally based on video content and video description. Please respond with short and concise answers. Avoid using unpronouncable punctuation or emojis. \\
\tt \\
\tt User: These are frames from a video. Your job is to explain why the video is humorous in 2-3 sentences as if you were explaning to a friend who doesn't get the joke yet. Respond with a 2-3 sentence explanation of the joke and how it relates to the video.\ \textbackslash n\textbackslash n \\
\tt \\
\tt Output format:\ \textbackslash n \\
\tt Explanation: \textless answer\textgreater\ \textbackslash n\textbackslash n \\
\bottomrule
\end{tabular}

\label{app_tab: prompt_for_video_exp}

\end{table}

\begin{table}[h!]
\centering
\caption{Prompt for video caption matching.}

\begin{tabular}{p{0.96\linewidth}}
\toprule
\tt System: You are a helpful AI assistant. You can analyze videos and answer questions about their content. Please only output in the specified format. No extra text. \\
\tt \\
\tt User: Along with the frames from the video. And \{question\}\ \textbackslash n \\
\tt Please respond with response with the option letter only.\ \textbackslash n\textbackslash n \\
\tt \\
\tt Output format:\ \textbackslash n \\
\bottomrule
\end{tabular}

\label{app_tab: prompt_for_video_caption_matching}

\end{table}

\begin{table}[h!]
\centering
\caption{Prompt for video with description QA.}

\begin{tabular}{p{0.96\linewidth}}
\toprule
\tt System: You are a helpful AI assistant. You can analyze videos and answer questions about their content. Respond with short and concise answers. Avoid using unpronouncable punctuation or emojis. \\
\tt \\
\tt User: You'll be given a description of the video. Based on this information, answer the following question: \{question\}\ \textbackslash n\textbackslash n \\
\tt \\
\tt Output format:\ \textbackslash n \\
\tt Answer: \textless answer\textgreater\ \textbackslash n\textbackslash n \\
\tt \\
\tt Video Description: \{video\_description\} \\
\bottomrule
\end{tabular}

\label{app_tab: prompt_for_video_dec_qa}

\end{table}

\begin{table}[h!]
\centering
\caption{Prompt for video with description explanation.}

\begin{tabular}{p{0.96\linewidth}}
\toprule
\tt System: You are a helpful AI assistant specialized in video understanding and humor analysis. You can explain jokes clearly and naturally based on video content and video description. Please respond with short and concise answers. Avoid using unpronouncable punctuation or emojis. \\
\tt \\
\tt User: You will also be given a description of the video. Your job is to explain why the video is humorous in 2-3 sentences as if you were explaning to a friend who doesn't get the joke yet. Respond with a 2-3 sentence explanation of the joke and how it relates to the video.\ \textbackslash n\textbackslash n \\
\tt \\
\tt Output format:\ \textbackslash n \\
\tt Explanation: \textless answer\textgreater\ \textbackslash n\textbackslash n \\
\tt \\
\tt Video Description: \{video\_description\} \\
\bottomrule
\end{tabular}

\label{app_tab: prompt_for_video_description_explanation}

\end{table}

\begin{table}[ht]
\centering
\caption{Prompt for writing captions of videos.}
\resizebox{\linewidth}{!}{%
\begin{tabular}{p{10cm}}
\toprule
\tt And I will provide a description of the video and a list of descriptive captions that break down what happens in it. \\
\tt Your task is to write a caption in one sentences from the video creator's perspective — something you would write to attract viwers.\\
\tt Requirements: \\
\tt Please ensure it is related to the video content. \\
\tt - Write as if you're sharing it with an audience (e.g., use 'this' or 'me' naturally). \\
\tt Output format: \\
\tt Caption: <caption> \\
\tt Video description: \{video\_description\} \\
\tt Descriptive captions: \{descriptive\_captions\}\\
\bottomrule
\end{tabular}
}
\label{app_tab: prompt_writing_caption_of_videos}
\end{table}

\begin{table}[h!]
\centering
\caption{Prompt for video with description caption matching.}

\begin{tabular}{p{0.96\linewidth}}
\toprule
\tt System: You are a helpful AI assistant. You can analyze videos and answer questions about their content. Please only output in the specified format. No extra text. \\
\tt \\
\tt User: You'll be given a description of the video. And \{question\}\textbackslash n Please respond with response with the option letter only.\textbackslash n\textbackslash n\\
\tt \\
\tt Output format:\textbackslash n\\
\tt Answer: \textless answer\textgreater \textbackslash n\textbackslash n\\ 
\tt Video Description: \{video\_description\} \\
\bottomrule
\end{tabular}

\label{app_tab: prompt_for_video_description_caption_matching}

\end{table}

\begin{table}[htbp]
\centering
\caption{Prompt for video with sound QA.}

\begin{tabular}{p{0.96\linewidth}}
\toprule
\tt System: You are a helpful AI assistant. You can analyze videos and answer questions about their content. Respond with short and concise answers. Avoid using unpronouncable punctuation or emojis. \\
\tt \\
\tt User: Here's a humorous video. Based on the its visual and audio information, answer the following question: \{question\}\ \textbackslash n\textbackslash n \\
\tt \\
\tt Output format:\ \textbackslash n \\
\tt Answer: \textless answer\textgreater\ \textbackslash n\textbackslash n \\
\bottomrule
\end{tabular}

\label{app_tab: prompt_for_video_with_sound_QA}

\end{table}

\begin{table}[htbp]
\centering
\caption{Video with sound explanation.}

\begin{tabular}{p{0.96\linewidth}}
\toprule
\tt System: You are a helpful AI assistant specialized in video understanding and humor analysis. You can explain jokes clearly and naturally based on video content and video description. Please respond with short and concise answers. Avoid using unpronouncable punctuation or emojis. \\
\tt \\
\tt User: Here's a humorous video. Your job is to explain why the video is humorous in 2-3 sentences as if you were explaning to a friend who doesn't get the joke yet. Respond with a 2-3 sentence explanation of the joke and how it relates to the video.\ \textbackslash n\textbackslash n \\
\tt \\
\tt Output format:\ \textbackslash n \\
\tt Explanation: \textless answer\textgreater\ \textbackslash n\textbackslash n \\
\bottomrule
\end{tabular}

\label{app_tab: prompt_for_video_with_sound_exp}

\end{table}

\begin{table}[t!]
\centering
\caption{Video with sound caption matching}

\begin{tabular}{p{0.96\linewidth}}
\toprule
\tt System: You are a helpful AI assistant. You can analyze videos and answer questions about their content. Please only output in the specified format. No extra text. \\
\tt \\
\tt User: Along with visual and audio information in the video. And \{question\}\ \textbackslash n \\
\tt Please respond with response with the option letter only.\ \textbackslash n\textbackslash n \\
\tt \\
\tt Output format:\ \textbackslash n \\
\tt Answer: \textless answer\textgreater\ \textbackslash n\textbackslash n \\
\bottomrule
\end{tabular}

\label{app_tab: prompt_for_video_with_sound_caption_matching}

\end{table}

\begin{table}[t!]
\centering
\caption{Prompt for video with background knowledge caption matching}
\resizebox{\linewidth}{!}{%
\begin{tabular}{p{10cm}}
\toprule
\tt System: You are a helpful AI assistant. You can analyze videos and answer questions about their content. Respond with short and concise answers. Avoid using unpronounceable punctuation or emojis. \\
\tt \\
\tt User: Here's a humorous video. You will be given background knowledge of the video. Based on its visual and audio information and the background knowledge, answer the following question: \{question\}\ \textbackslash n \\
\tt Please respond with the option letter only.\ \textbackslash n\textbackslash n \\
\tt \\
\tt Output format:\ \textbackslash n \\
\tt Answer: \textless answer\textgreater\ \textbackslash n\textbackslash n \\
\tt Background Knowledge: \{background\_knowledge\}\\
\bottomrule
\end{tabular}

\label{app_tab: prompt_for_video_with_background_knowledge_caption_matching}
}
\end{table}

\end{document}